\documentclass[a4paper,fleqn]{cas-dc}

\usepackage[numbers]{natbib}
\usepackage{xurl}
\usepackage[expansion=false]{microtype}
\newcommand{\costgrouprules}{\cmidrule(lr){2-3}\cmidrule(lr){4-5}}

\begin{document}
\let\WriteBookmarks\relax
\def\floatpagepagefraction{1}
\def\textpagefraction{.001}

\shorttitle{Image Rotation Angle Estimation: Five Circular-Aware Methods Across Sixteen Architectures}    

\shortauthors{M. Woehrer}  

\title [mode = title]{Image Rotation Angle Estimation: Five Circular-Aware Methods Across Sixteen Architectures}

\author[1,2]{Maximilian Woehrer}[orcid=0000-0001-8536-4900]

\cormark[1]

\ead{maximilian.woehrer@univie.ac.at}

\credit{Conceptualization, Methodology, Software, Validation, Formal analysis, Investigation, Data curation, Writing -- original draft, Writing -- review \& editing, Visualization}

\affiliation[1]{organization={Research Group Software Architecture, Faculty of Computer Science, University of Vienna},
            city={1090 Vienna},
            country={Austria}}

\affiliation[2]{organization={MAXI solutions e.U.},
            city={1170 Vienna},
            country={Austria}}

\cortext[1]{Corresponding author}

\begin{abstract}
Automatic image rotation estimation is a key preprocessing step in many vision pipelines, yet challenging because angles have circular topology, creating boundary discontinuities that hinder standard regression.
We present a comprehensive study of five circular-aware methods for global orientation estimation: three regression based (direct angle with circular loss, unit vector, phase-shifting coder) and two classification based (angular binning, circular Gaussian distribution).
Using transfer learning from ImageNet-pretrained models, we evaluate them across sixteen modern architectures and two data regimes of natural images, small-scale DRC-D and large-scale COCO.
Classification approaches are the stronger family, attaining the lowest error on thirteen of the sixteen DRC-D backbones.
The decisive ingredient is supervision that encodes angular neighborhood structure: plain cross-entropy lacks it and fails intermittently, while circular smooth labels (CSL) eliminate every observed training failure and reach the lowest DRC-D error, 0.81\textdegree{} mean absolute error (MAE).
On COCO, the common rotate-and-crop protocol leaks the applied angle through the rotated JPEG block grid, a shortcut that the classification heads learned and the regression heads largely did not.
Re-encoding every training view after rotation suppresses it.
Under this trace-free protocol and on re-encoded test views, classification again leads on all three backbones tested, and the circular Gaussian distribution (CGD) with ConvNeXt~V2 Base reaches 3.76\textdegree{} on COCO~2017 (three-run mean), against 6.09\textdegree{} for the strongest prior model on the same views.
CSL and CGD are also the cheapest methods to train.
\end{abstract}

\begin{graphicalabstract}
\includegraphics{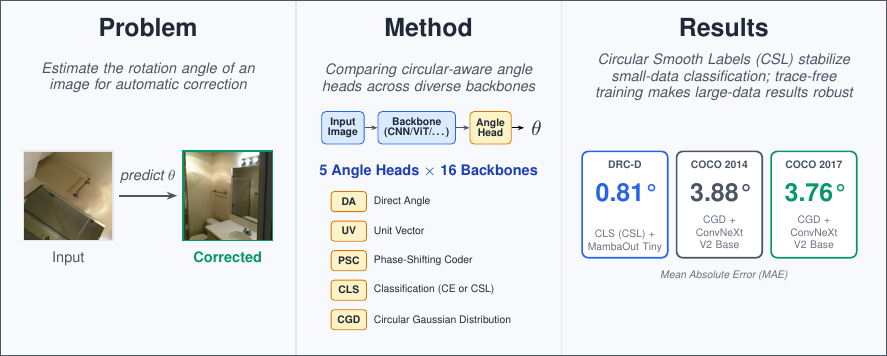}
\end{graphicalabstract}

\begin{highlights}
\item Five circular-aware angle methods, regression vs.\ classification, on 16 backbones
\item Classification beats regression on 13/16 small-data and all 3 large-data backbones
\item Circular smooth labels cure small-data classification instability, DRC-D MAE 0.81\textdegree{}
\item Rotate-and-crop leaks the angle via the JPEG grid; re-encoding views suppresses it
\item Trace-free CGD + ConvNeXt V2 Base: 3.76\textdegree{} MAE on COCO 2017, prior model 6.09\textdegree{}
\end{highlights}


\maketitle

\section{Introduction}

Image rotation estimation is a common preprocessing step in computer vision pipelines, ensuring that images are properly oriented before further analysis. This task can be approached in two ways: discrete classification, which predicts cardinal rotations (e.g., 0\textdegree{}, 90\textdegree{}, 180\textdegree{}, 270\textdegree{}), or continuous prediction across the full 360\textdegree{} range.

Continuous angle prediction is particularly challenging because angles have circular topology, creating two fundamental problems. First, the same physical orientation can be expressed by infinitely many numerical values: 0\textdegree{}, 360\textdegree{}, and 720\textdegree{} all describe identical rotations. Second, the circular boundary creates artificial discontinuities where physically similar angles appear numerically distant: 359\textdegree{} and 1\textdegree{} are only 2\textdegree{} apart in reality but appear 358\textdegree{} apart numerically.
These circular properties cause significant problems for standard deep learning approaches. When neural networks predict angles as scalar values using typical L1 or L2 losses, they treat equivalent orientations as different targets and compute artificially large errors at the circular boundary~\cite{zhouContinuityRotationRepresentations2019,levinsonAnalysisSVDDeep}. This severely hampers learning and model performance unless the circular nature of the data is explicitly addressed.

Existing approaches to global image rotation estimation each address this circularity in isolation: early CNN systems used standard regression losses that ignore the circular topology entirely~\cite{fischerImageOrientationEstimation2015}, later work introduced a circular loss but kept a scalar output~\cite{majiDeepImageOrientation2020}, and the richer circular-aware output representations developed in oriented object detection and pose estimation have not been systematically applied to this task. Each individual scheme was proposed and evaluated on a single architecture, so it is unknown whether its reported advantage stems from the angle representation, the backbone, or their interaction.
Furthermore, no comprehensive study has compared different circular-aware methods across modern neural architectures. With the evolution from traditional CNNs to Vision Transformers and efficient architectures, practitioners lack clear guidance on which methods work best with which models. This gap motivates the present study: rather than proposing yet another output head, we hold the training pipeline fixed and systematically vary both the circular-aware method and the backbone.

To address these limitations, we present a systematic evaluation of five circular-aware methods in two families: the regression approaches direct angle regression with circular loss (DA), unit vector regression (UV), and phase-shifting coder (PSC); and the classification approaches classification via angular binning (CLS) and circular Gaussian distribution (CGD). We test these methods across sixteen modern architectures and different data regimes to identify the best combinations and provide practical guidance for method selection. %We additionally evaluate a circularly smoothed label variant of the classification head that resolves the training instabilities we observe.

Such a comparison is only meaningful if the models estimate the angle from the image content. Training and test views for this task are produced by rotating and cropping upright photographs, and we show that this common procedure leaves a processing trace, the rotated JPEG block grid of the source photograph, from which a network can read the applied angle without regard to the content. We identify which heads learn this shortcut and in which data regime, suppress it with a single re-encoding step, and carry out the large-data comparison under the resulting trace-free protocol (Section~\ref{sec:traces}).

The remainder of this paper is organized as follows. Section~\ref{sec:related} reviews related work. Section~\ref{sec:methods} describes the five circular-aware methods. Section~\ref{sec:experiments} details the datasets and experimental setup. Section~\ref{sec:results} presents the results together with ablations and controls. Section~\ref{sec:analysis} discusses the findings, Section~\ref{sec:guidance} distills practical guidance, and Section~\ref{sec:limitations} states limitations and future work. Section~\ref{sec:conclusion} concludes.

\section{Related Work}\label{sec:related}

Traditional rotation estimation methods relied on hand-crafted features such as color distributions, edge orientations, and texture descriptors combined with classical machine learning approaches.
The introduction of deep convolutional networks marked a paradigm shift, with Fischer et al.~\cite{fischerImageOrientationEstimation2015} establishing a foundational CNN-based approach that achieved 20.97\textdegree{} MAE on COCO 2014 images, though it used standard regression losses that treat angular data as scalar values. Maji and Bose~\cite{majiDeepImageOrientation2020} achieved 8.38\textdegree{} MAE using an Xception architecture with specialized angular loss functions, and their code repository has since been updated with a Vision Transformer model claiming 6.5\textdegree{} MAE, though this later result remains unpublished. Transfer learning from ImageNet-pretrained models has further improved results across rotation estimation tasks~\cite{amjoudTransferLearningAutomatic2022,joshiAutomaticPhotoOrientation2017}, and Dong et al.~\cite{dongBlindDetectionCircular2022} combine ensemble transfer regression with fused HOG features for blind full-range rotation estimation, though their scalar regression target retains the boundary-discontinuity problem. These developments highlight both the potential of modern architectures and the persistent challenges posed by circular data topology.

The boundary discontinuity at 0\textdegree{}/360\textdegree{} has driven a range of circular-aware methods, largely developed in oriented object detection~\cite{xiaDOTALargeScaleDataset2018,dingLearningRoITransformer2019,chengOrientedCenterNetRotated2023} and pose estimation~\cite{pavlloQuaterNetQuaternionbasedRecurrent2018}; Cheng et al.~\cite{chengOrientedCenterNetRotated2023}, for example, sidestep direct angle regression in rotated detection by re-parameterizing boxes through corner-point offsets and splitting the regression loss into horizontal and rotated parts, illustrating how detection pipelines engineer around the discontinuity rather than solving it in the angle representation itself. Theoretical analyses have shown that low-dimensional rotation representations are inherently discontinuous, motivating boundary-free parameterizations~\cite{zhouContinuityRotationRepresentations2019,levinsonAnalysisSVDDeep}. Direct regression with circular losses~\cite{majiDeepImageOrientation2020} computes minimum angular distances but remains vulnerable to gradient conflicts. Classification approaches discretize the angular space, with techniques like Circular Smooth Labels~\cite{yangArbitraryOrientedObjectDetection2020} and Dense Coded Labels~\cite{yangDenseLabelEncoding2021} incorporating periodic smoothing and Gray coding for boundary handling. Probabilistic methods model orientations as distributions, either continuous ones such as von Mises mixtures~\cite{prokudinDeepDirectionalStatistics2018} and the Bingham distribution~\cite{gilitschenskiDeepOrientationUncertainty2020} or discretized ones such as Circular Gaussian Distributions~\cite{xuRotatedObjectDetection2023}, enabling uncertainty quantification. Phase-Shifting Coders~\cite{yuPhaseShiftingCoderPredicting2023} provide continuous, boundary-free representations through multiple cosine components with different phase offsets. Recent work has further sought theoretically guaranteed continuous representations~\cite{xiaoTheoreticallyAchievingContinuous2024}, reflecting ongoing interest in this problem.

While Xu et al.~\cite{xuImageOrientationEstimation2024} provide an overview of classification and regression approaches for orientation estimation, these methods have not been systematically compared for global image rotation estimation across modern architectures.

\section{Circular-Aware Methods}\label{sec:methods}

Building on the foundations established in related work, we now present five circular-aware methods that address the boundary discontinuity problem through different paradigms. We organize these methods into regression and classification approaches, each addressing the circular topology challenge through distinct strategies. We provide a brief overview of each method, referring to the cited literature for more details. Throughout, $\theta$ denotes the ground-truth rotation angle and $\hat{\theta}$ the prediction. All five methods consume the same input features: the globally pooled feature vector extracted by the (pretrained) backbone from the rotated image. They differ in the output head and its parameterization of the angle, in the loss used for training, and in how predictions are decoded back to $\hat{\theta}$.

\subsection{Regression Methods}

Regression methods predict continuous angle values while carefully handling the circular topology of angular data.
These methods address the boundary discontinuity at 0\textdegree{}/360\textdegree{} through specialized loss functions, continuous parameterizations, or encoding schemes that maintain differentiability for gradient-based optimization.

\subsubsection{Direct Angle with Circular Loss}

The most intuitive approach predicts orientation angles directly through a single output neuron. While conceptually straightforward, this method requires careful loss function design to handle the circular nature of angular data. Traditional loss functions treat angles near boundaries as numerically distant despite their geometric proximity, leading to optimization difficulties.

Circular-aware loss functions address this limitation by incorporating angular distance into the objective function. Following Maji and Bose~\cite{majiDeepImageOrientation2020}, we use a circular mean absolute error that computes the loss based on the shorter angular distance $\min(|\hat{\theta} - \theta|, 360^\circ - |\hat{\theta} - \theta|)$. This formulation ensures that predictions near angular boundaries receive appropriate gradient signals during training, as the loss correctly treats 1\textdegree{} and 359\textdegree{} as only 2\textdegree{} apart rather than 358\textdegree{} apart.

The direct angle approach remains vulnerable to gradient conflicts when network predictions span the 0\textdegree{}/360\textdegree{} boundary during training, but the circular loss formulation improves convergence compared to standard regression losses.

Our implementation uses a single regression head with circular MAE loss, providing a deliberately minimal baseline for comparison with more sophisticated circular representations. We note that this formulation faces two distinct obstacles: the circular loss itself is non-convex in the scalar output (its sawtooth structure changes gradient sign at the wrap-around), and the one-dimensional output representation is inherently discontinuous for circular targets~\cite{zhouContinuityRotationRepresentations2019}. %Its expected failure therefore reflects the representation as much as the loss, a distinction we probe with a two-output ablation in Section~\ref{sec:results}.

\subsubsection{Unit Vector Approach}

Unit vector representation avoids angular boundaries entirely by parameterizing orientations as points on the unit circle. The network outputs two values representing cosine and sine components: $[\cos(\theta), \sin(\theta)]$. This representation naturally handles the circular topology since unit vectors provide continuous coverage of the angular space without discontinuities. Tsai et al.~\cite{tsaiPreciseOrientationEstimation2024} demonstrated the effectiveness of this unit vector coding approach for precise orientation estimation in rotated object detection.

Rather than explicitly normalizing outputs to unit length during forward passes, our implementation uses regularization terms that encourage unit magnitude while preserving gradient flow, following Pavllo et al.~\cite{pavlloQuaterNetQuaternionbasedRecurrent2018}. The total loss combines MAE between predicted and target unit vectors with a regularization term $\lambda (\|v\| - 1)^2$, where $\lambda = 0.01$ penalizes deviation from unit magnitude. This strategy balances mathematical correctness with optimization stability.

Decoding predicted unit vectors to angles uses the two-argument arctangent function: $\hat{\theta} = \text{atan2}(v_{\sin}, v_{\cos})$, which correctly handles all quadrants and provides unique angle recovery across the full 360\textdegree{} range.

\subsubsection{Phase-Shifting Coder}

Phase-shifting approaches encode angles through multiple cosine components with different phase offsets, providing a continuous and boundary-free representation~\cite{yuPhaseShiftingCoderPredicting2023}. The encoding uses $M$ phase-shifted terms: $m_n = \cos(\omega\theta + 2\pi n/M)$ for $n = 0, 1, \ldots, M-1$. This parameterization distributes angular information across multiple outputs while maintaining differentiability.

The phase-shifting formulation addresses boundary discontinuity by ensuring smooth variation across the angular space. Since cosine functions are periodic and continuous, the encoded representation avoids the sharp transitions that plague direct angle prediction. 

Decoding uses $\hat{\theta} = -(1/\omega)\,\allowbreak\text{atan2}(S_s, S_c)$, where $S_s = \sum_n m_n\sin(2\pi n/M)$, $S_c = \sum_n m_n\cos(2\pi n/M)$; as in the unit vector approach, the two-argument arctangent resolves all quadrants.
The network is trained to predict the phase-shifted cosine values directly using MAE loss between predicted and target PSC representations.

Our implementation uses three phases with unit frequency ($\omega = 1$), providing a balance between representation capacity and computational efficiency.

\subsection{Classification Methods}

Classification methods discretize the 360\textdegree{} angular space into bins, framing orientation estimation as a multi-class prediction problem. This paradigm leverages established classification frameworks and often provides stable optimization dynamics. The key challenge is handling the circular relationship between bins, particularly the boundary between 0\textdegree{} and 360\textdegree{}, which can be addressed through specialized loss functions and probabilistic formulations.

\subsubsection{Classification}

Classification approaches discretize the 360\textdegree{} angular space into $N$ bins, treating rotation estimation as a standard multi-class problem. Each bin represents a range of angles, with bin centers serving as discrete angle representatives. This discretization enables the use of well-established classification techniques and loss functions. This approach has been applied in automatic photo rotation estimation~\cite{joshiAutomaticPhotoOrientation2017} and transfer learning scenarios~\cite{amjoudTransferLearningAutomatic2022}, though those works discretize into only four coarse classes (0\textdegree{}, 90\textdegree{}, 180\textdegree{}, 270\textdegree{}); we extend the idea to fine-grained 360-bin classification for continuous rotation estimation.

The classification approach offers several practical advantages, including straightforward integration with existing classification frameworks. Note, however, that discretization removes the wrap-around discontinuity only from the target encoding: cross-entropy itself treats the bins as unordered classes, so the angular adjacency of neighboring bins (including bins 0 and 359 across the boundary) is not represented in the objective. The predicted angle is decoded as the center of the highest-probability bin (argmax of softmax outputs), yielding angle resolution equal to the bin width. Our baseline implementation employs 360 bins (1\textdegree{} per bin) with standard cross-entropy loss for training.

Because plain cross-entropy encodes no angular neighborhood structure, we additionally implement and evaluate the classification head with Circular Smooth Labels (CSL)~\cite{yangArbitraryOrientedObjectDetection2020}: the one-hot target is replaced by a soft label distribution obtained from a circular window function centered on the true bin, so that bins adjacent to the true angle (including across the 0\textdegree{}/359\textdegree{} boundary) receive partial probability mass. We use the Gaussian window with radius 6\textdegree{} and $\sigma = 2^\circ$, the configuration reported as strongest in the original work, train with cross-entropy against these soft labels, and decode by argmax exactly as in the plain variant; architecture, parameter count, and inference are unchanged, making CSL a pure change of supervision. We adopt CSL as the representative circular label-smoothing remedy; Dense Coded Labels (DCL)~\cite{yangDenseLabelEncoding2021} primarily target the code-length efficiency of long one-hot vectors rather than the neighborhood structure itself and are therefore not evaluated here, though an implementation is included in our released code for completeness.

\subsubsection{Circular Gaussian Distribution}

The Circular Gaussian Distribution (CGD) approach represents angles as probability distributions over discretized angular bins. Rather than predicting point estimates, the network outputs a probability distribution that captures both the predicted angle and associated uncertainty. Ground truth targets are encoded as Gaussian distributions centered at the true angle. This approach was introduced by Xu et al.~\cite{xuRotatedObjectDetection2023} for rotated object detection to address boundary discontinuity issues in oriented bounding box regression.

Target encoding creates smooth probability distributions using circular distances with probability proportional to $\exp(-d^2(\phi_i,\theta)/(2\sigma^2))$ where $d$ is the circular distance between bin center $\phi_i$ and target angle $\theta$. Training uses Kullback-Leibler divergence between predicted and target distributions, while decoding uses argmax over the predicted distribution to identify the peak bin.

The original formulation uses 180 bins for the $[-90^\circ, 90^\circ)$ range of oriented bounding boxes; we adapt it to 360 bins covering the full $[0^\circ, 360^\circ)$ range for global image rotation, keeping $\sigma = 6^\circ$ as recommended by Xu et al.

\section{Experiments}
\label{sec:experiments}

\subsection{Datasets and Study Design}
Our experiments are organized in two dataset regimes, and the two datasets were chosen to serve exactly these roles.

The small-data regime uses the DRC-D dataset, introduced for content-preserving rotation correction~\cite{nieDeepRotationCorrection2023}. We use the ground truth images from this dataset, which contain a large diversity of natural scenes with clear upright indicators, avoiding ambiguous orientations such as aerial views or abstract scenes that lack clear orientation cues. Its manageable size (1,474 unique training, 535 unique testing images) makes an exhaustive grid affordable: the sixteen architectures crossed with the six method columns reported throughout (the five methods, with the classification head trained under both plain cross-entropy and circular smooth labels) give 96 configurations, each trained five times, for 480 training runs.

The large-data regime uses the COCO 2014 dataset~\cite{linMicrosoftCOCOCommon2015} (${\sim}$83K training images), as used by Fischer et al.~\cite{fischerImageOrientationEstimation2015} and Maji and Bose~\cite{majiDeepImageOrientation2020}. Here every method is trained on three representative backbones, which tests whether the small-data findings survive a 56-fold increase in training data; the best configuration of that grid is additionally trained on COCO 2017 (${\sim}$117K images) to probe the effect of further scale, with three training runs on each of the two datasets. COCO 2017 re-partitions the images of COCO 2014, so the 915 of our 1{,}030 test images that fall into its training split were removed from it, leaving 117{,}372 training images; the COCO 2014 training split contains none of the test images, which stem from its validation split. COCO also enables direct comparison with prior work: we use the same 1,030 test images as Fischer et al., comparing against their reported MAE while noting that we cannot replicate their exact rotation methodology due to undisclosed test rotation parameters. For Maji and Bose's method~\cite{majiDeepImageOrientation2020}, whose exact test set of 1,000 filtered COCO images is unknown, we evaluate their released model on the Fischer test set, on the same rotated views as our own models, to maintain consistent comparison conditions across all works.

\subsection{Rotation Pipeline}
During training and testing, we apply synthetic rotations to images and handle the resulting geometric transformations using the largest rotated rectangle approach. Rotation angles are measured counter-clockwise, and the network is trained to predict the angle that was applied to the upright source image. During training, a fresh angle is sampled uniformly at random from $[0^\circ, 360^\circ)$ for every image in every epoch; validation and test angles are drawn once from the same uniform distribution under a fixed seed and then held constant. The processing order is identical for all backbones: the source image is first rotated (with the canvas expanded to retain all content), then center-cropped to the largest axis-aligned rectangle inscribed in the rotated content, and only afterwards resized and normalized by the backbone-specific input transform. This method crops the maximum area from rotated images while avoiding black border artifacts that a network could otherwise exploit as a direct cue to the applied angle. Alternative approaches exist, such as Follmann and Böttger's~\cite{follmannRotationallyinvariantConvolutionModule2018} circular cropping method that uses a circle with radius equal to half the image size, with black borders smoothed using a Gaussian kernel (size 5 pixels), but we choose the rotated rectangle approach for its area maximization properties.

Avoiding border artifacts is not sufficient, however. Photographs are stored as JPEG files, and decoding one leaves the $8\times 8$ block grid of the codec in its pixels. Rotating the decoded image turns this grid together with the content, so every rotated view carries the applied angle modulo 90\textdegree{} in a faint periodic pattern that is independent of what the image shows. Section~\ref{sec:traces} shows that networks trained on COCO learn to read it. For the large-data regime we therefore change the pipeline in two ways: after rotation and cropping, every training and validation view is re-encoded once as a JPEG at a quality drawn uniformly from 40 to 90, and the interpolation kernel of the rotation, bilinear in the conventional protocol, is drawn at random (bilinear, bicubic or Lanczos). The re-encoding overlays a grid that is aligned with the image frame, as any camera or web pipeline produces, and that carries no information about the angle. We refer to this as the trace-free protocol and to the pipeline without the re-encoding step as the conventional protocol; the name refers to the rotated block grid, and Section~\ref{sec:limitations} states what the protocol does not guarantee. The large-data grid was trained under both protocols: its models of the conventional protocol serve the analysis of Section~\ref{sec:traces}, and all other large-data results are those of the trace-free protocol. DRC-D is trained and evaluated under the conventional protocol; Section~\ref{sec:traces} examines its effect at that scale and finds that no DRC-D model collapses on re-encoded test views and that the family ordering is retained there.

\subsection{Architectures and Heads}
\begin{figure}
\centering
\includegraphics[width=\columnwidth]{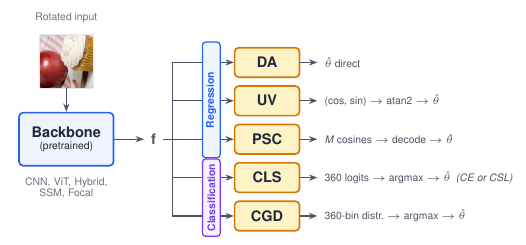}
\caption{Model architecture. A pretrained backbone extracts features from the rotated input, which are passed to one of five circular-aware angle heads to predict the rotation angle~$\hat{\theta}$; the classification head is trained with either plain cross-entropy (CE) or circular smooth labels (CSL).}
\label{fig:pipeline}
\end{figure}

Our approach follows established transfer learning methodology similar to Amjoud and Amrouch~\cite{amjoudTransferLearningAutomatic2022}, leveraging ImageNet-pretrained~\cite{russakovskyImageNetLargeScale2015} models as feature extractors. Each architecture serves as a backbone with its final classifier replaced by task-specific heads optimized for the respective circular-aware formulation. This strategy enables efficient comparison across methods while benefiting from rich pretrained representations. Figure~\ref{fig:pipeline} illustrates the model architecture.

We select eight architecture families that span the major design paradigms in modern visual recognition: pure transformers (ViT~\cite{dosovitskiyImageWorth16x162021}, Swin~\cite{liuSwinTransformerHierarchical2021}), pure convolutions (ConvNeXt V2~\cite{wooConvNeXtV2Codesigning2023}, EfficientNetV2~\cite{tanEfficientNetV2SmallerModels}), hybrid designs (EfficientViT~\cite{liuEfficientViTMemoryEfficient2023}, EdgeNeXt~\cite{maazEdgeNeXtEfficientlyAmalgamated2022}), focal attention (FocalNet~\cite{yangFocalModulationNetworks}), and state-space-inspired gating (MambaOut~\cite{yuMambaOutWeReally}). For each family we include a small and a large variant to disentangle the effect of model capacity from architectural inductive bias. The complete set of 16 backbones is listed in Table~\ref{tab:overview_all}. The timm library~\cite{wightmanPyTorchImageModels2019} provides consistent backbone handling across different architectures. For regression approaches (direct angle, unit vector, PSC), we use a regression head with progressive dimensionality reduction (from backbone features through hidden layers of decreasing size), layer normalization, ReLU activation, and dropout regularization. Classification and CGD approaches use standard linear layers to produce the required number of output logits. Backbone capacities span 1.2M parameters (EdgeNeXt XX-Small) to 87.7M (ConvNeXt V2 Base, FocalNet Base); the output heads add comparatively little, with the linear heads used by CLS (plain and CSL variant) and CGD contributing 0.06--1.1M parameters and the MLP regression head used by DA, UV, and PSC contributing 0.02--6.2M, both scaling with the backbone's feature width, so the comparison is dominated by the shared backbone cost.

\subsection{Training and Evaluation Protocol}
All experiments were run on a single NVIDIA A100 GPU. We use PyTorch Lightning (mixed precision for the DRC-D grid, 32-bit precision for the COCO experiments), AdamW~\cite{loshchilovDecoupledWeightDecay2019} with square-root batch-size learning-rate scaling (to adapt each architecture's base learning rate to the training batch size), ReduceLROnPlateau learning-rate scheduling, and early stopping with patience of 15 epochs (on COCO, an improvement must exceed 0.1\textdegree{} of validation MAE to reset the patience counter), training for up to 1{,}000 epochs. We use a batch size of 16; on a small dataset, larger batches yield too few gradient steps per epoch relative to the epoch-based patience window, preventing adequate fine-tuning of pretrained weights. No data augmentations are applied during training to avoid interfering with orientation learning; the only exceptions are the two steps of the trace-free protocol, the re-encoding and the random interpolation kernel, which are applied to every view on COCO. All backbones share the identical rotation-and-crop pipeline described above and differ only in the resolution of the final input transform, which follows each backbone's published configuration (for example $224 \times 224$ for most models): the view is resized with bicubic interpolation so that its shorter side matches this resolution, its central square is taken without the additional crop margin of the standard evaluation transform, and the ImageNet normalization statistics are applied. We use fixed random seeds: each run's seed controls the train/validation split (10\% validation, drawn from the training images only) and generates consistent validation rotation angles, while training applies random rotations [0\textdegree{}, 360\textdegree{}) each epoch. A seed does not make a training run repeatable, since weight initialization and batch order are not seeded and, in the runs reported here, the split depended on the order in which the file system listed the images (the released code lists them in sorted order); reproducibility of training is therefore statistical (Section~\ref{sec:limitations}). The test evaluation is exactly repeatable: the order of the test images, which pairs them with the seeded test angles, is pinned in the released code. The validation split is used solely for early stopping, checkpoint selection, and learning-rate scheduling; all reported results are computed on the held-out test sets (535 DRC-D or 1{,}030 COCO images), which are fixed, disjoint image sets that are never part of any training or validation split. Because each of the five runs uses a different seed, the five runs also cover five different random train/validation partitions, so the reported means additionally reflect split variability in the spirit of repeated random sub-sampling validation. On DRC-D, a single fixed test seed is shared across all configurations to ensure directly comparable evaluation; the reported variability reflects training randomness alone. On COCO, we additionally average over five test seeds to account for test-rotation variability, and we evaluate every model on two forms of the same test views: re-encoded once as a JPEG at quality 90 after rotation and cropping, and without re-encoding. The re-encoded form is our primary condition, because it corresponds to how photographs are stored and its final save largely overwrites the rotated grid; the form without re-encoding is the test condition of the conventional protocol and is reported in Appendix~\ref{app:traces}. This asymmetric protocol, five training runs per configuration on DRC-D versus a single run per cell on COCO (three runs for the best configuration), is motivated by the training-cost gap between the two regimes, quantified in Section~\ref{sec:cost}; a sensitivity check on the validation-split size is reported in Section~\ref{sec:ablations}.

\subsection{Metrics}
We define the circular distance between predicted angle $\hat{\theta}$ and true angle $\theta$ as the minimum angular separation: $d(\hat{\theta},\theta) = \min(|\hat{\theta}-\theta|, 360^\circ-|\hat{\theta}-\theta|)$. This distance metric forms the foundation for evaluating all methods in our study. We report MAE, RMSE, and median error from circular distances, together with Acc@$k$\textdegree{} (the share of errors of at most $k$\textdegree{}) and AUC@$k$\textdegree{} (the area under the curve of this share against the threshold from 0 to $k$\textdegree{}, normalized to one) for $k \in \{2,5,10\}$, and the 90th and 95th percentile errors (P90, P95) to characterize tail behavior. All statistics are computed over the whole test set. For DRC-D we report the MAE, the accuracies and the AUC values; the median, RMSE and percentile errors require the predictions of a complete test run, which are available only for the best run of each configuration. Standard deviations are population standard deviations.

\section{Results}\label{sec:results}

Section~\ref{sec:drcd} presents the comparison in the small-data regime of the study design (Section~\ref{sec:experiments}). Section~\ref{sec:traces} examines the processing traces of the rotation pipeline, which determine how the large-data results have to be obtained, and Sections~\ref{sec:coco_matrix}--\ref{sec:coco2017} present the large-data comparison and the effect of scaling the training set. Section~\ref{sec:error_analysis} analyzes the error structure of the best large-data configuration. Section~\ref{sec:ablations} collects the ablations and controls, each labeled with the dataset it was run on, and Section~\ref{sec:cost} reports training cost. Throughout, we compare the five methods, with the classification head appearing under both supervision variants (CE and CSL).

\subsection{Small-Data Regime: DRC-D}\label{sec:drcd}

Table~\ref{tab:overview_all} shows results across all architecture-method combinations, grouped into the regression and classification families of Section~\ref{sec:methods}.
Read by family, the two paradigms separate clearly: a classification head attains the lowest MAE on 13 of the 16 backbones, regression winning only on ConvNeXt~V2 Atto (UV), EfficientNetV2-RW~T (PSC) and ViT-Tiny (UV), and the best classification cell (0.81\textdegree{}) is well ahead of the best regression cell (1.47\textdegree{}). The advantage depends heavily on the loss encoding angular neighborhood structure, however. By column median the ordering is CLS (CSL) 1.86\textdegree{}, CGD 2.16\textdegree{}, UV 2.55\textdegree{}, PSC 2.70\textdegree{}, CLS (CE) 3.30\textdegree{} and DA 15.42\textdegree{}, so plain cross-entropy, which treats the 360 bins as unordered classes, ranks behind both stable regression methods despite holding the best single classification configuration among the base methods.
Among the five base methods (the columns other than CLS (CSL)), CLS with EfficientViT-B3 achieves the best mean MAE of 1.23\textdegree{}, but exhibits training instability on several backbone combinations where some runs fail to converge: MambaOut Tiny (35.94\textdegree{}), MambaOut Base (55.15\textdegree{}), Swin Base (20.33\textdegree{}), ConvNeXt V2 Base (20.71\textdegree{}), and EfficientNetV2-RW~T (29.39\textdegree{}); the CSL variant introduced below eliminates exactly these failures. CGD is the strongest of the five overall, achieving the lowest MAE among them on 9 of 16 architectures; like UV and PSC, and unlike DA and plain CLS, it converged in every run on every backbone; CGD with MambaOut Base reaches 1.24\textdegree{}, virtually matching the best CLS result. These two configurations are statistically indistinguishable: a paired t-test across the five seed-matched runs (runs with the same seed share the train/validation split) yields $p = 0.97$ (Wilcoxon signed-rank $p = 1.0$), and the bootstrap 95\% confidence interval of the per-seed difference, $[-0.79^\circ, +0.79^\circ]$, comfortably contains zero. The practically relevant distinction between CLS and CGD is therefore not mean accuracy but convergence reliability. PSC and UV achieve competitive results on the ConvNeXt~V2 family, with PSC+ConvNeXt~V2 Base being the most stable configuration in the whole grid (std = 0.16\textdegree{}). Direct-angle regression remains unstable across all backbones despite the circular loss.

\begin{figure}[pos=tbp]
\centering
\includegraphics[width=0.9\columnwidth]{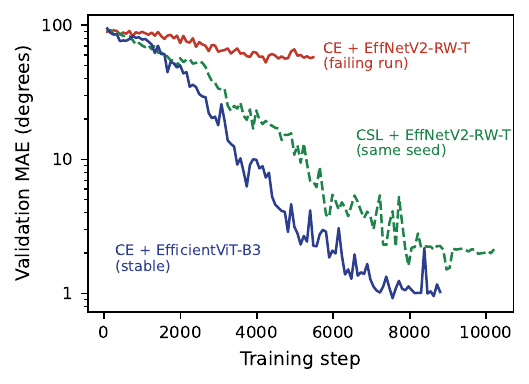}
\caption{Validation MAE during training on DRC-D (log scale), for representative diagnostic runs. A failing plain cross-entropy (CE) run on EfficientNetV2-RW~T plateaus at high error, while the CSL variant on the same backbone and the same seed converges; CE on a compatible backbone (EfficientViT-B3) is shown for reference.}
\label{fig:loss_curves}
\end{figure}

\medskip\noindent\textbf{Circular smooth labels resolve the CLS instability.}
To test whether the CLS training failures stem from the missing circular neighborhood structure in plain cross-entropy, we trained the CSL variant of the classification head on all sixteen backbones; the results form the CLS (CSL) column of Table~\ref{tab:overview_all}. CSL eliminates the failure mode entirely: all 80 runs converge (worst single run 10.0\textdegree{}), with no catastrophic outcomes on any backbone, including all five where plain cross-entropy failed. Beyond stabilizing training, CSL improves over plain cross-entropy on 14 of 16 backbones and yields the best configuration on 8 of the 16 architectures, matching CGD's robustness while reaching lower error. Its best configuration, CSL with MambaOut Tiny, reaches 0.81\textdegree{}~$\pm$~0.31\textdegree{}, the best DRC-D result in this study, on a backbone where plain cross-entropy failed (35.94\textdegree{}); the same holds for MambaOut Base, whose plain-cross-entropy cell fails most severely (55.15\textdegree{}) and which CSL brings to 1.05\textdegree{}. This supports the interpretation that the instability stems from the absence of angular neighborhood structure in the loss rather than by properties of the backbone alone (Section~\ref{sec:analysis}). Statistically, CSL+MambaOut Tiny is nominally the strongest configuration (bootstrap 95\% confidence interval of its mean $[0.54^\circ, 1.08^\circ]$), but five paired runs do not separate it from the best plain-CLS and CGD configurations: the per-seed differences are $-0.41^\circ$ and $-0.43^\circ$ with bootstrap intervals $[-0.91^\circ, +0.01^\circ]$ and $[-1.02^\circ, +0.07^\circ]$ and paired t-tests at $p = 0.19$ and $p = 0.25$, so we describe it as the consistently best-performing configuration rather than a statistically separated one. Figure~\ref{fig:loss_curves} illustrates the mechanism at the level of individual runs: a failing plain-cross-entropy run plateaus at high validation error for its entire training, while the CSL variant on the same backbone and seed converges.

\begin{table*}[t]
\centering
\caption{DRC-D test performance across methods and backbones: the five circular-aware methods, with the classification head trained under both supervision variants, plain cross-entropy (CE) and circular smooth labels (CSL). MAE\textdegree{}\ values are means (standard deviation in parentheses) across five independent training runs. Best method per architecture in bold; column-best marked with $^*$; overall best underlined. Right-hand metrics are 5-run means for the best method per architecture (bold). $^\dagger$Several CLS (CE) configurations exhibit high variance due to intermittent training instability; reported means include all five runs, and the CLS (CSL) column shows the remedy. The best CLS (CE) and best CGD configurations are statistically indistinguishable (paired t-test $p = 0.97$; Section~\ref{sec:drcd}). All models are trained and tested under the conventional protocol; Section~\ref{sec:traces} reports their best runs on re-encoded views.}
\label{tab:overview_all}
\setlength{\tabcolsep}{1.5pt}
\renewcommand{\arraystretch}{0.85}
\footnotesize
\resizebox{\textwidth}{!}{%
\begin{tabular}{l|ccc|ccc|cc}
\toprule
\multirow{2}{*}{Architecture} & \multicolumn{3}{c|}{Regression (MAE\textdegree{})} & \multicolumn{3}{c|}{Classification (MAE\textdegree{})} & \multicolumn{2}{c}{Best method metrics (5-run means)} \\
 & DA & UV & PSC & CLS (CE) & CLS (CSL) & CGD & Acc (2\textdegree{}/5\textdegree{}/10\textdegree{}) & AUC (2\textdegree{}/5\textdegree{}/10\textdegree{}) \\
\midrule
ConvNeXt V2 Atto & 7.88(2.11)$^*$ & \textbf{2.69(0.41)} & 2.83(0.64) & 3.04(1.55) & 3.92(2.03) & 3.46(1.54) & 0.53/0.89/0.98 & 0.28/0.57/0.76 \\
ConvNeXt V2 Base & 61.42(8.16) & 1.54(0.36)$^*$ & 1.47(0.16)$^*$ & 20.71(33.63)$^\dagger$ & \textbf{0.93(0.47)} & 2.22(1.23) & 0.98/1.00/1.00 & 0.70/0.88/0.94 \\
EdgeNeXt XX-Small & 20.16(2.70) & 4.22(1.48) & 3.54(0.64) & 3.29(1.28) & 2.33(1.33) & \textbf{2.10(0.61)} & 0.78/0.99/0.99 & 0.43/0.73/0.86 \\
EdgeNeXt Base & 14.78(3.02) & 1.88(0.53) & 2.30(0.82) & 1.77(1.74) & \textbf{1.65(0.91)} & 2.34(1.08) & 0.94/0.99/0.99 & 0.62/0.84/0.92 \\
EfficientNetV2-RW~T & 12.79(1.10) & 3.16(0.44) & \textbf{2.83(0.49)} & 29.39(30.06)$^\dagger$ & 3.37(0.81) & 3.14(1.01) & 0.51/0.87/0.98 & 0.26/0.54/0.74 \\
EfficientNetV2-RW~M & 11.42(2.02) & 2.83(0.85) & 2.50(0.96) & 2.68(0.87) & \textbf{1.90(0.33)} & 3.47(0.50) & 0.81/0.99/0.99 & 0.46/0.75/0.87 \\
EfficientViT-B0 & 15.68(4.06) & 5.87(2.01) & 4.94(0.87) & 7.46(1.13) & 6.20(2.18) & \textbf{4.87(0.52)} & 0.56/0.92/0.97 & 0.29/0.59/0.77 \\
EfficientViT-B3 & 8.25(1.84) & 2.41(0.35) & 2.74(1.53) & \textbf{1.23(0.61)}$^*$ & 1.59(0.98) & 1.93(0.33) & 0.90/0.99/1.00 & 0.55/0.80/0.90 \\
FocalNet Tiny LRF & 12.59(3.94) & 2.34(0.60) & 2.66(0.82) & 3.31(0.92) & \textbf{1.82(0.89)} & 2.28(0.63) & 0.91/0.99/0.99 & 0.58/0.81/0.90 \\
FocalNet Base LRF & 11.32(2.69) & 2.28(0.59) & 2.35(0.37) & 3.18(1.64) & 2.23(2.01) & \textbf{1.81(0.59)} & 0.77/0.99/1.00 & 0.42/0.73/0.86 \\
MambaOut Tiny & 60.23(10.47) & 2.20(0.81) & 2.09(0.70) & 35.94(40.68)$^\dagger$ & \underline{\textbf{0.81(0.31)}}$^*$ & 1.33(0.22) & 0.99/1.00/1.00 & 0.71/0.88/0.94 \\
MambaOut Base & 70.07(8.65) & 1.99(0.38) & 2.57(1.21) & 55.15(42.98)$^\dagger$ & \textbf{1.05(0.60)} & 1.24(0.53)$^*$ & 0.98/1.00/1.00 & 0.69/0.87/0.93 \\
Swin Tiny & 74.23(6.96) & 3.30(1.39) & 2.21(0.62) & 2.42(0.98) & \textbf{1.48(0.86)} & 2.03(0.90) & 0.94/0.99/0.99 & 0.62/0.84/0.92 \\
Swin Base & 78.87(10.09) & 2.31(0.65) & 2.89(0.99) & 20.33(34.64)$^\dagger$ & \textbf{1.02(0.43)} & 1.39(0.39) & 0.96/1.00/1.00 & 0.65/0.85/0.93 \\
ViT-Tiny & 18.39(1.76) & \textbf{4.16(0.76)} & 4.17(0.22) & 6.85(0.74) & 6.76(1.96) & 5.69(1.56) & 0.43/0.81/0.96 & 0.22/0.48/0.69 \\
ViT-Base & 15.16(2.83) & 2.89(0.65) & 2.80(0.58) & 2.97(1.80) & 2.16(0.94) & \textbf{2.10(0.49)} & 0.70/0.98/1.00 & 0.37/0.69/0.84 \\
\bottomrule
\end{tabular}}
\end{table*}

\medskip\noindent\textbf{Backbone trends.} Among the five base methods, EfficientViT-B3 is the standout backbone for classification (1.23\textdegree{}); once smoothed labels are used, MambaOut Tiny becomes classification's strongest partner (0.81\textdegree{} under CSL), closely followed by ConvNeXt~V2 Base (0.93\textdegree{}) and Swin Base (1.02\textdegree{}). MambaOut architectures also pair best with CGD, producing the top-2 CGD results (1.24\textdegree{}\ for MambaOut Base and 1.33\textdegree{}\ for MambaOut Tiny). ConvNeXt V2 Base pairs most reliably with PSC and UV among the regression methods. The plain-CLS instability (training failures on some architectures) affects small and large backbones alike (MambaOut Tiny and Base, Swin Base, ConvNeXt V2 Base, EfficientNetV2-RW~T), so it does not follow model capacity, and the CSL result points to the loss as the cause (Section~\ref{sec:analysis}). Direct-angle regression performs far worse than every stable alternative on all backbones and collapses outright on many, consistent with analyses showing that smoothing the loss alone does not remove boundary issues~\cite{xuRethinkingBoundaryDiscontinuity2024,yuBoundaryDiscontinuityAngle2024}; as discussed in Section~\ref{sec:methods}, its one-dimensional output representation is a contributing factor alongside the non-convex circular loss.

\subsection{Processing Traces in the Rotation Pipeline}\label{sec:traces}

Before comparing the methods at scale, we examine whether models trained on rotated and cropped photographs estimate the angle from the image content. There is reason to doubt it. Compression and resampling leave periodic traces in an image: the $8\times 8$ block grid of JPEG coding can be detected and its quantization estimated from the decoded pixels~\cite{fanIdentificationBitmapCompression2003}, and interpolation introduces periodic correlations that reveal resampling~\cite{popescuExposingDigitalForgeries2005}. Networks readily exploit such low-level regularities when they predict the training target, a behaviour known as shortcut learning~\cite{geirhosShortcutLearningDeep2020}. In self-supervised learning, chromatic aberration was found to solve a patch-position pretext task~\cite{doerschUnsupervisedVisualRepresentation2015}, and Gidaris et al.~\cite{gidarisUnsupervisedRepresentationLearning2018} restrict their rotation pretext task to multiples of 90\textdegree{} to avoid the low-level artifacts of other geometric transformations.

The conventional protocol of Section~\ref{sec:experiments} offers a regularity of this kind, the rotated block grid, and the classification models trained under it on COCO use it. Table~\ref{tab:traces} lists every large-data model trained under the conventional protocol on the test views without re-encoding and on the same views saved once at JPEG quality 90. CGD with MambaOut Base trained on COCO 2017, for example, goes from 2.84\textdegree{} to 24.45\textdegree{}, and on re-encoded views about three quarters of its predictions lie within 2\textdegree{} of a multiple of 90\textdegree{}. This sensitivity was first reported, for a publicly released model of ours, by ORTUS AI~\cite{ortusaiRightWayUp2026}, who conjectured the JPEG block grid as its cause.

\medskip\noindent\textbf{The cue is the rotated block grid.}
Two controls identify what these models read. The first is a content-free probe: isotropic noise textures, which contain no orientation, are saved as JPEG, rotated and cropped by the pipeline, and shown to the model. Because the grid has four-fold symmetry, a model that reads it recovers the angle modulo 90\textdegree{}; by chance, 11.1\% of predictions fall within 5\textdegree{} of it. The model named above places 12.9\% of its predictions there when the textures carry no grid, and 42.1\%, 78.1\% and 96.1\% when they were saved at quality 90, 75 and 50. The second control puts content and grid in conflict: on test photographs whose own grid is suppressed by a blur, a grid is stamped at an offset of 20\textdegree{} from the content. The two conventional-protocol CGD models we tested then follow the grid on 97\% of the images, yet still choose the correct quadrant on about 90\%. Content thus decides the coarse orientation and the grid the fine angle, which also explains the sub-degree median errors of these models.

\medskip\noindent\textbf{Mainly the binned heads learned it.}
Table~\ref{tab:traces} separates the two families sharply. The UV and PSC heads change by 0.2\textdegree{} to 1.2\textdegree{} under re-encoding, and direct-angle regression by at most 0.3\textdegree{}, whereas the three binned heads (CE, CSL and CGD) lose between 4.5\textdegree{} and 20.3\textdegree{} on COCO 2014, most on ConvNeXt~V2 Base, and up to 21.6\textdegree{} on COCO 2017. On re-encoded views, every UV and PSC model of the conventional protocol is ahead of every classification model on the same backbone. A plausible reason is the target: one-degree bins reward any cue that sharpens an estimate below a degree, whereas a regression loss on a two- or three-dimensional code changes little once a prediction is within a degree or two. The margin by which classification led regression at scale under the conventional protocol was therefore inflated by the cue, and the large-data comparison has to be repeated without it.

\medskip\noindent\textbf{The small-data regime is not affected in the same way.}
The DRC-D source images are JPEG files as well, so the cue is present there too. We evaluated the best run of every configuration of Table~\ref{tab:overview_all} on re-encoded views (96 checkpoints; only the run with the lowest test error of each configuration was retained); Table~\ref{tab:drcd_reenc} gives the outcome by head. No model collapses. UV and PSC are unaffected, whereas plain cross-entropy and CSL lose part of their fine precision, 0.3\textdegree{} of MAE on average (CGD: 0.1\textdegree{}), and the three binned heads lose 0.6\textdegree{} to 1.1\textdegree{} in ten of their 48 configurations, among them the best run of CSL with MambaOut Tiny (0.48\textdegree{} to 1.08\textdegree{}), the configuration behind the best five-run mean of 0.81\textdegree{}. The family ordering is retained nevertheless, although re-encoding takes from the classification heads a cue that was available to them in training and leaves the regression heads as they are: among these best runs, a classification head has the lowest MAE on 14 of the 16 backbones without re-encoding and on 13 with it, and CSL remains the strongest head by median over the backbones. On the content-free probe the regression heads stay at chance on average for every grid strength; the binned heads do so for textures without a grid and with a quality-90 grid, and respond only to stronger grids. A control in which DRC-D was re-encoded at quality 75 before training, which strengthens the grid, gives the same picture for CGD on two backbones (0.92\textdegree{} versus 0.90\textdegree{} and 0.90\textdegree{} versus 0.91\textdegree{} without and with re-encoding of the test views, probe at chance). With 1{,}474 training images the cue sharpens the fine angle of some classification models without deciding the comparison; with 83K it carries the fine angle. The method comparison of Table~\ref{tab:overview_all} therefore stands, while the sub-degree values of its best classification cells should be read as values under the conventional protocol.

\begin{figure}[pos=tbp]
\centering
\includegraphics[width=0.9\columnwidth]{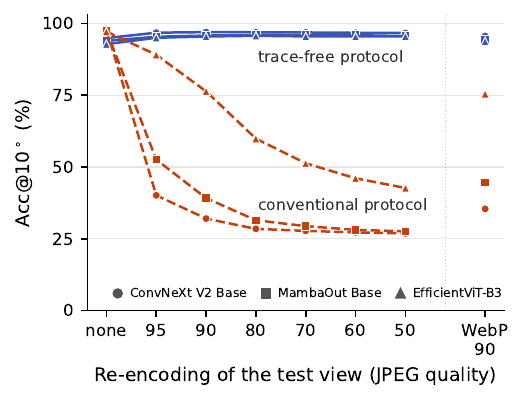}
\caption{Acc@10\textdegree{} of CGD on the COCO test views without re-encoding (``none'') and after one save at JPEG quality 95 to 50 or as WebP at quality 90, for the models trained on COCO 2014 under the trace-free protocol (solid) and under the conventional protocol (dashed); means over five test seeds.}
\label{fig:quality}
\end{figure}

\medskip\noindent\textbf{Re-encoding after rotation suppresses the cue.}
Under the trace-free protocol of Section~\ref{sec:experiments}, a model never sees a rotated grid without a frame-aligned one on top, and the results of Sections~\ref{sec:coco_matrix} and~\ref{sec:coco2017} are obtained this way. We verify every trace-free model with four checks (Appendix~\ref{app:traces}), three of them with pass criteria that were fixed before the evaluations were run: the content-free probe, two controls on the test photographs that measure how much precision a model draws from a rotated source grid and from the interpolation pattern of the rotation, and a cue-conflict test. Under cue conflict, all converged trace-free models with a structured head (every head except direct-angle regression) follow the content, the offset grid attracting at most 6\% of their predictions. All pass both photograph controls: a rotated source grid changes their Acc@2\textdegree{} by $-1.3$ to $+0.6$ points, whereas two conventional-protocol models, tested with a lossless final save as in their own protocol, gain 23 and 29 points. On the content-free probe, 16 of the 18 grid cells pass every condition; CSL with ConvNeXt~V2 Base and plain cross-entropy with MambaOut Base miss the share criterion, each COCO 2017 run misses it in one condition by at most 1.5 points (Appendix~\ref{app:traces}), and these models are marked in Table~\ref{tab:coco_all}. Trace-free models are also stable across codec settings (Figure~\ref{fig:quality}): for CGD on the three backbones, Acc@10\textdegree{} stays above 95\% from JPEG quality 95 down to 50 and above 93\% for WebP, a codec these models never saw in training. The conventional-protocol models lose accuracy with the mildest re-encoding, falling to between 40\% and 89\% at quality 95, and WebP has the same effect, so the collapse is not tied to particular JPEG settings.

\begin{table}[pos=tbp]
\centering
\caption{Large-data models trained under the conventional protocol: MAE\textdegree{}\ on test views without re-encoding, the test condition of that protocol, and on the same views re-encoded once at JPEG quality 90, means across five test seeds. The regression heads UV and PSC are nearly unaffected, the binned heads are not.}
\label{tab:traces}
\setlength{\tabcolsep}{3pt}
\renewcommand{\arraystretch}{0.92}
\footnotesize
\resizebox{\columnwidth}{!}{%
\begin{tabular}{llcccccc}
\toprule
Backbone & Test views & DA & UV & PSC & CLS (CE) & CLS (CSL) & CGD \\
\midrule
\multicolumn{8}{l}{\textit{COCO 2014}} \\
ConvNeXt V2 Base & no re-encoding & 63.51 & 4.45 & 4.88 & 3.35 & 4.28 & 3.13 \\
 & re-encoded & 63.55 & 4.86 & 5.14 & 19.39 & 19.73 & 23.45 \\
EfficientViT-B3 & no re-encoding & 11.44 & 4.96 & 5.04 & 4.50 & 4.24 & 3.88 \\
 & re-encoded & 11.74 & 5.21 & 5.24 & 9.42 & 11.01 & 9.90 \\
MambaOut Base & no re-encoding & 59.55 & 5.20 & 4.68 & 4.83 & 3.80 & 3.79 \\
 & re-encoded & 59.69 & 5.60 & 5.86 & 10.52 & 8.27 & 21.48 \\
\midrule
\multicolumn{8}{l}{\textit{COCO 2017, CGD only}} \\
ConvNeXt V2 Base & no re-encoding & -- & -- & -- & -- & -- & 2.97 \\
 & re-encoded & -- & -- & -- & -- & -- & 24.13 \\
EfficientViT-B3 & no re-encoding & -- & -- & -- & -- & -- & 4.00 \\
 & re-encoded & -- & -- & -- & -- & -- & 10.61 \\
MambaOut Base & no re-encoding & -- & -- & -- & -- & -- & 2.84 \\
 & re-encoded & -- & -- & -- & -- & -- & 24.45 \\
\bottomrule
\end{tabular}}
\end{table}

\begin{table}[pos=tbp]
\centering
\caption{DRC-D models on re-encoded test views, by head: the best run of each of the 96 configurations of Table~\ref{tab:overview_all} (16 backbones per head), evaluated without re-encoding and re-encoded once at JPEG quality 90. Median MAE\textdegree{} over the backbones; change of the MAE\textdegree{} under re-encoding as mean over the backbones, largest rise, and number of backbones with a rise of at least 0.5\textdegree{}; share of predictions within 5\textdegree{} of the grid angle on content-free textures with a quality-50 grid (chance 11.1\%), as mean and maximum over the backbones.}
\label{tab:drcd_reenc}
\setlength{\tabcolsep}{3pt}
\renewcommand{\arraystretch}{0.95}
\footnotesize
\resizebox{\columnwidth}{!}{%
\begin{tabular}{lccccccc}
\toprule
 & \multicolumn{2}{c}{Median MAE\textdegree{}} & \multicolumn{3}{c}{Change of MAE\textdegree{}} & \multicolumn{2}{c}{Probe share (\%)} \\
\cmidrule(lr){2-3}\cmidrule(lr){4-6}\cmidrule(lr){7-8}
Head & no re-enc. & re-enc. & mean & largest & $\geq 0.5$\textdegree{} & mean & max \\
\midrule
DA & 12.03 & 12.19 & $+$0.11 & $+$0.43 & 0 & 11.6 & 15.0 \\
UV & 1.73 & 1.83 & $+$0.02 & $+$0.19 & 0 & 11.7 & 15.2 \\
PSC & 1.69 & 1.73 & $+$0.04 & $+$0.13 & 0 & 11.4 & 15.5 \\
CLS (CE) & 1.27 & 1.58 & $+$0.30 & $+$1.12 & 3 & 18.1 & 41.2 \\
CLS (CSL) & 0.84 & 1.22 & $+$0.33 & $+$0.99 & 6 & 14.6 & 27.0 \\
CGD & 1.36 & 1.40 & $+$0.09 & $+$0.63 & 1 & 12.8 & 18.0 \\
\bottomrule
\end{tabular}}
\end{table}

\subsection{Large-Data Regime: COCO 2014}\label{sec:coco_matrix}

The DRC-D grid identifies, for each method, the backbone it performs best with; this leaves open whether a method ranking established on 1{,}474 images reflects the methods themselves or these dataset-specific pairings. We therefore train every method on COCO 2014 with each of three backbones from different architecture families (EfficientViT-B3, MambaOut Base, ConvNeXt~V2 Base), and compare against two historical baselines: Net-360~\cite{fischerImageOrientationEstimation2015} and OAD-360~\cite{majiDeepImageOrientation2020}, whose released model we re-evaluate on our test views. Table~\ref{tab:coco_all} collects the COCO results; they are obtained under the trace-free protocol and reported on re-encoded test views, and Appendix~\ref{app:traces} lists the same models on views without re-encoding. Regarding comparability, only the OAD-360 row is directly comparable to our methods, since we evaluate their released Vision Transformer model on identical test views; the Net-360 value and OAD-360's originally published 8.38\textdegree{} (obtained with their earlier Xception model) are literature-reported numbers obtained under rotation and test-set protocols that are not fully specified and are included for historical context only. Each grid cell is a single training run, motivated by the training-cost asymmetry between the regimes (Section~\ref{sec:cost}). One run failed visibly: plain cross-entropy on ConvNeXt~V2 Base remained at chance level (90.3\textdegree{}), and following the rule stated in Section~\ref{sec:limitations} it was repeated with a different seed; the repeat, which stayed at chance level for more than thirty epochs before it converged, is the reported run.

The results indicate that the family-level pattern established on DRC-D holds at scale. On every backbone the best classification method beats the best regression method, by 3.52\textdegree{} versus 4.80\textdegree{} on ConvNeXt~V2 Base, 4.51\textdegree{} versus 5.27\textdegree{} on EfficientViT-B3, and 4.09\textdegree{} versus 4.69\textdegree{} on MambaOut Base; by column median the ordering is CSL 4.09\textdegree{}, CE 4.58\textdegree{}, CGD 4.75\textdegree{}, PSC 4.80\textdegree{}, UV 5.05\textdegree{} and DA 64.77\textdegree{}. The margins are smaller than on DRC-D, and they depend on the test condition: on views without re-encoding, which lie slightly outside the training distribution of these models, the errors of the structured methods are higher by 0.2\textdegree{} to 2.6\textdegree{}, classification stays ahead on ConvNeXt~V2 Base and EfficientViT-B3, and PSC leads on MambaOut Base (4.90\textdegree{} versus 4.98\textdegree{} for CSL; Table~\ref{tab:coco_noreenc}). Within the classification family no variant dominates. CGD is best on ConvNeXt~V2 Base (3.52\textdegree{}), plain cross-entropy on EfficientViT-B3 (4.51\textdegree{}) and CSL on MambaOut Base (4.09\textdegree{}), each ahead of the runner-up by 0.08\textdegree{} to 0.49\textdegree{}, which is within the variation between training runs of one configuration (Section~\ref{sec:coco2017}). Direct-angle regression remains far behind, reaching 11.57\textdegree{} on its best backbone and failing on the other two (64.77\textdegree{} and 67.85\textdegree{}, against 90\textdegree{} at chance). On re-encoded views, every structured method on every backbone outperforms OAD-360 (6.09\textdegree{}), which in turn beats the early CNN baseline Net-360 (20.97\textdegree{}). Two things do not transfer from the small-data regime, however. The CSL variant's dominance does not: its decisive small-data advantage reduces to parity with the other classification heads here (Section~\ref{sec:analysis} examines why). Neither does the optimal pairing: the best configuration at scale is CGD with ConvNeXt~V2 Base, whereas CGD's best partner on DRC-D is MambaOut Base, suggesting that a backbone chosen on the smaller dataset is not necessarily the right choice at scale.

OAD-360 itself reaches 6.10\textdegree{} without re-encoding and 6.09\textdegree{} on re-encoded views, close to the 6.5\textdegree{} its authors state for this model, and it is insensitive to the re-encoding, as expected for a model that was not trained on rotated JPEG grids. It is behind our structured heads on typical images and in the upper percentiles alike, with a median of 2.08\textdegree{} against 0.33\textdegree{}--1.93\textdegree{} and a P95 of 14.2\textdegree{} against 6.4\textdegree{}--11.5\textdegree{}. The RMSE, which is dominated by the rare errors near 90\textdegree{} and 180\textdegree{}, is of the same size (20.1\textdegree{} against 14.1\textdegree{}--21.4\textdegree{}).

\subsection{Scaling the Training Set: COCO 2017}\label{sec:coco2017}

To probe the effect of further scale, the configuration with the lowest error in the grid, CGD with ConvNeXt~V2 Base (lowest by 0.08\textdegree{}), was trained on the larger COCO 2017 train set (${\sim}$117K images). So that this comparison does not rest on single runs, the configuration was trained three times on each dataset with different seeds (lower block of Table~\ref{tab:coco_all}). On COCO 2014 the three runs reach 3.52\textdegree{} (the run of the grid), 3.87\textdegree{} and 4.24\textdegree{} (mean 3.88\textdegree{}, standard deviation 0.29\textdegree{}); on COCO 2017 they reach 3.70\textdegree{}, 3.77\textdegree{} and 3.81\textdegree{} (mean 3.76\textdegree{}, standard deviation 0.05\textdegree{}). The larger training set thus leaves the mean error unchanged within the run-to-run variation of the smaller one. Its three runs agree far more closely, although three runs per dataset do not establish a difference in variability, and precision on typical images is slightly higher: Acc@2\textdegree{} rises from 0.68 to 0.71 and the median error falls from 1.28\textdegree{} to 1.21\textdegree{}. The COCO 2017 mean of 3.76\textdegree{} is the headline large-data result of this study; the lower value of 3.52\textdegree{} in the grid is a single run of a configuration whose three-run mean is 3.88\textdegree{}.

The three COCO 2014 runs also show how finely the single-run cells of the grid can be read: runs of one configuration differ by up to 0.72\textdegree{}, more than the margin of the best classification variant over the runner-up on any backbone.

\begin{table*}[t]
\centering
\caption{COCO evaluation (1{,}030 test images) under the trace-free protocol, on test views re-encoded once at JPEG quality 90, organised as in Table~\ref{tab:overview_all}, with the median, RMSE and percentile errors in addition; the same models on views without re-encoding are listed in Table~\ref{tab:coco_noreenc}. Upper block: prior work on COCO 2014, with OAD-360 evaluated on the same test views. Middle block: every method trained on each of three representative backbones on COCO 2014; MAE\textdegree{}\ values are means (standard deviation in parentheses) across five test seeds, from a single training run per cell (see Section~\ref{sec:limitations} for how finely single-run values should be read). Best method per backbone in bold; the right-hand metrics are for the per-backbone best method. Lower block: the best configuration trained three times on COCO 2014 and on COCO 2017; MAE\textdegree{}\ values are means (standard deviation in parentheses) across the three training runs, and the COCO 2017 mean (3.76\textdegree{}, underlined) is the lowest mean over training runs and the headline COCO result of this study. $^\ddagger$Misses the share criterion of the content-free probe (Section~\ref{sec:traces}, Appendix~\ref{app:traces}).}
\label{tab:coco_all}
\setlength{\tabcolsep}{3pt}
\renewcommand{\arraystretch}{0.85}
\footnotesize
\resizebox{\textwidth}{!}{%
\begin{tabular}{l|ccc|ccc|ccccc}
\toprule
\multirow{2}{*}{Architecture} & \multicolumn{3}{c|}{Regression (MAE\textdegree{})} & \multicolumn{3}{c|}{Classification (MAE\textdegree{})} & \multicolumn{5}{c}{Best method metrics} \\
 & DA & UV & PSC & CLS (CE) & CLS (CSL) & CGD & Med\textdegree{} & RMSE\textdegree{} & Acc (2\textdegree{}/5\textdegree{}/10\textdegree{}) & AUC (2\textdegree{}/5\textdegree{}/10\textdegree{}) & P90/P95\textdegree{} \\
\midrule
\multicolumn{12}{l}{\textit{COCO 2014, prior work}} \\
Net-360~\cite{fischerImageOrientationEstimation2015} (AlexNet-like) & \multicolumn{6}{c|}{20.97} & -- & -- & --/--/-- & --/--/-- & --/-- \\
OAD-360~\cite{majiDeepImageOrientation2020} (ViT) & \multicolumn{6}{c|}{6.09(0.73)} & 2.08 & 20.09 & 0.49/0.81/0.93 & 0.26/0.51/0.70 & 7.93/14.19 \\
\midrule
\multicolumn{12}{l}{\textit{COCO 2014 (${\sim}$83K train images), ours, one training run per cell}} \\
ConvNeXt V2 Base & 64.77(0.48) & 4.97(0.37) & 4.80(0.39) & 4.91(0.09) & 3.60(0.34)$^\ddagger$ & \textbf{3.52(0.28)} & 1.30 & 14.12 & 0.67/0.92/0.97 & 0.38/0.66/0.81 & 4.34/6.42 \\
EfficientViT-B3 & 11.57(0.56) & 5.30(0.23) & 5.27(0.34) & \textbf{4.51(0.23)} & 4.74(0.31) & 4.75(0.17) & 1.27 & 18.81 & 0.67/0.91/0.96 & 0.38/0.65/0.80 & 4.65/8.43 \\
MambaOut Base & 67.85(0.61) & 5.05(0.41) & 4.69(0.11) & 4.58(0.36)$^\ddagger$ & \textbf{4.09(0.34)} & 4.80(0.30) & 0.79 & 18.39 & 0.80/0.93/0.96 & 0.52/0.74/0.84 & 3.72/7.51 \\
\midrule
\multicolumn{12}{l}{\textit{CGD with ConvNeXt V2 Base, three training runs per dataset}} \\
COCO 2014 (${\sim}$83K train images) & -- & -- & -- & -- & -- & 3.88(0.29) & 1.28 & 16.21 & 0.68/0.92/0.97 & 0.39/0.66/0.81 & 4.20/6.61 \\
COCO 2017 (${\sim}$117K train images) & -- & -- & -- & -- & -- & \underline{3.76(0.05)}$^\ddagger$ & 1.21 & 16.81 & 0.71/0.94/0.97 & 0.40/0.68/0.82 & 3.83/5.74 \\
\bottomrule
\end{tabular}}
\end{table*}

\subsection{Error Analysis}\label{sec:error_analysis}

\begin{figure*}[width=\textwidth,pos=tp]
\centering
\setlength{\tabcolsep}{1pt}
\begin{tabular}{ccccccccccccc}
\rotatebox{90}{\footnotesize\textbf{Input}} &
\includegraphics[width=0.077\textwidth]{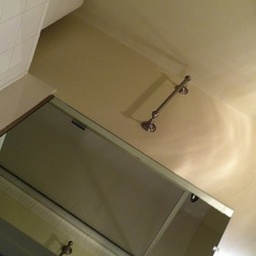} &
\includegraphics[width=0.077\textwidth]{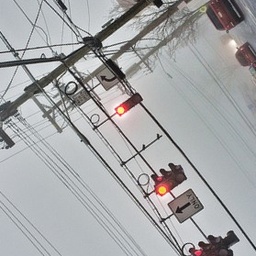} &
\includegraphics[width=0.077\textwidth]{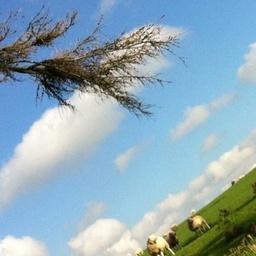} &
\includegraphics[width=0.077\textwidth]{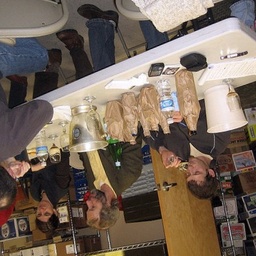} &
\includegraphics[width=0.077\textwidth]{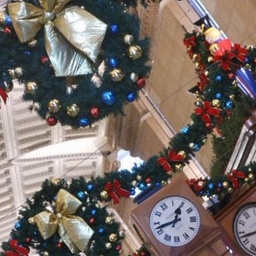} &
\includegraphics[width=0.077\textwidth]{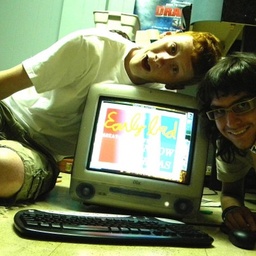} &
\includegraphics[width=0.077\textwidth]{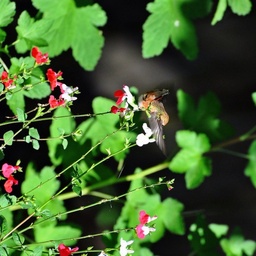} &
\includegraphics[width=0.077\textwidth]{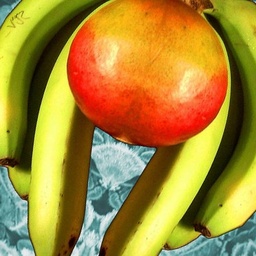} &
\includegraphics[width=0.077\textwidth]{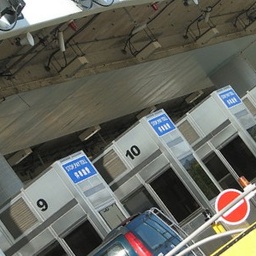} &
\includegraphics[width=0.077\textwidth]{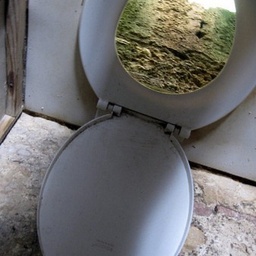} &
\includegraphics[width=0.077\textwidth]{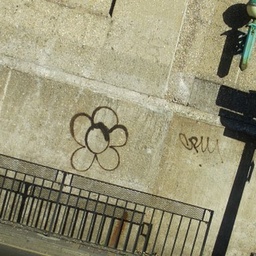} &
\includegraphics[width=0.077\textwidth]{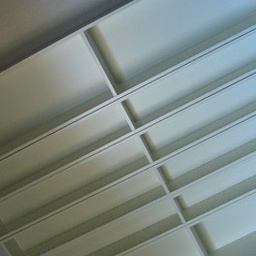} \\[1pt]
\rotatebox{90}{\footnotesize\textbf{CGD}} &
\includegraphics[width=0.077\textwidth]{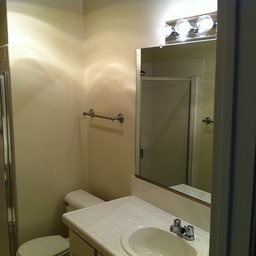} &
\includegraphics[width=0.077\textwidth]{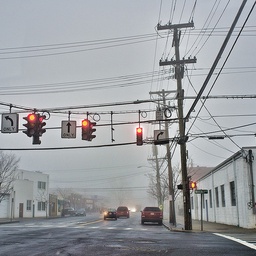} &
\includegraphics[width=0.077\textwidth]{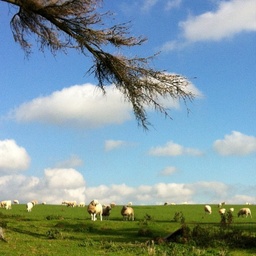} &
\includegraphics[width=0.077\textwidth]{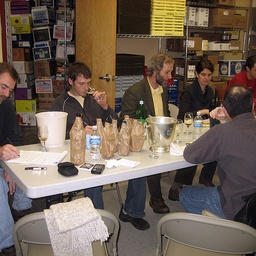} &
\includegraphics[width=0.077\textwidth]{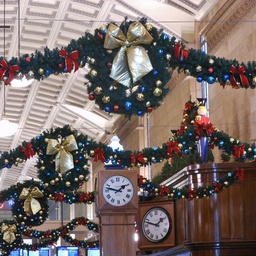} &
\includegraphics[width=0.077\textwidth]{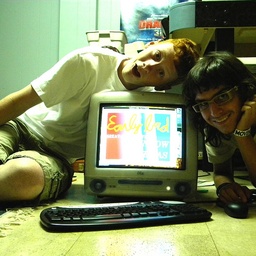} &
\includegraphics[width=0.077\textwidth]{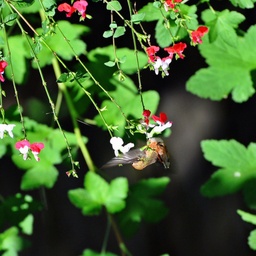} &
\includegraphics[width=0.077\textwidth]{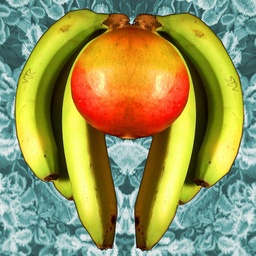} &
\includegraphics[width=0.077\textwidth]{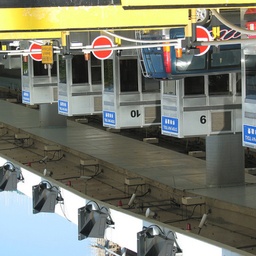} &
\includegraphics[width=0.077\textwidth]{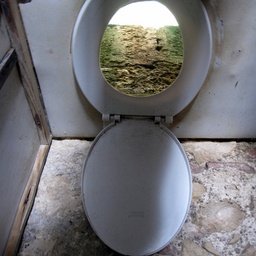} &
\includegraphics[width=0.077\textwidth]{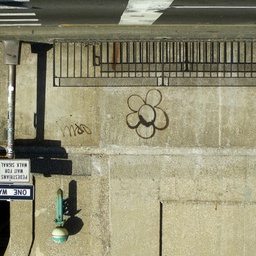} &
\includegraphics[width=0.077\textwidth]{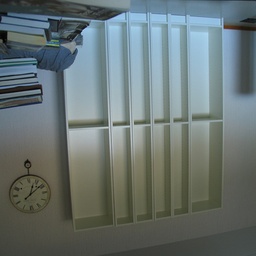} \\[1pt]
\rotatebox{90}{\footnotesize\textbf{GT}} &
\includegraphics[width=0.077\textwidth]{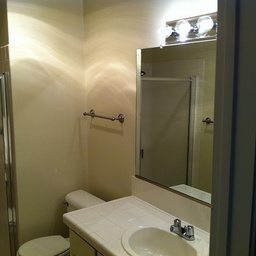} &
\includegraphics[width=0.077\textwidth]{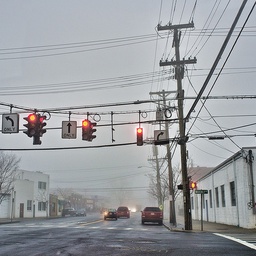} &
\includegraphics[width=0.077\textwidth]{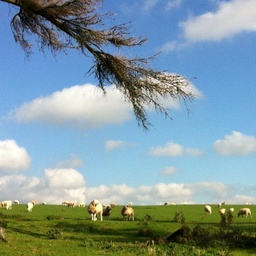} &
\includegraphics[width=0.077\textwidth]{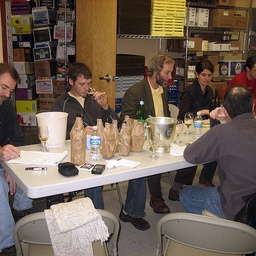} &
\includegraphics[width=0.077\textwidth]{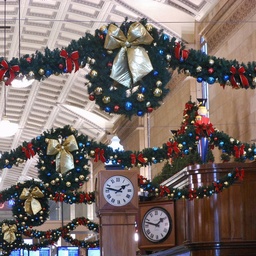} &
\includegraphics[width=0.077\textwidth]{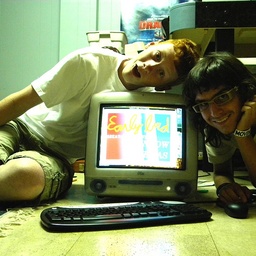} &
\includegraphics[width=0.077\textwidth]{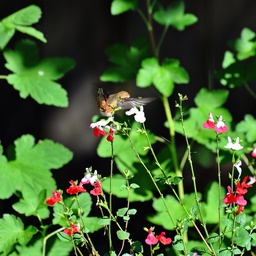} &
\includegraphics[width=0.077\textwidth]{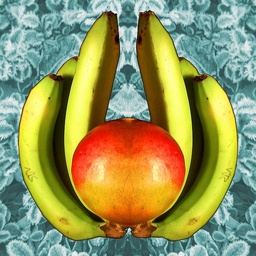} &
\includegraphics[width=0.077\textwidth]{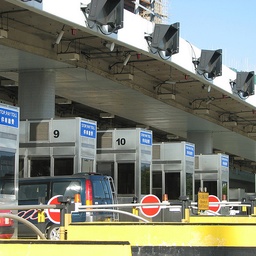} &
\includegraphics[width=0.077\textwidth]{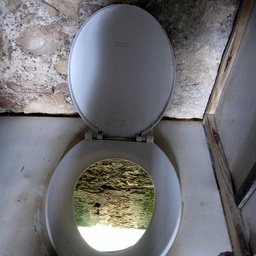} &
\includegraphics[width=0.077\textwidth]{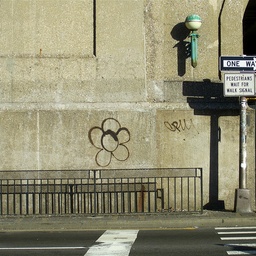} &
\includegraphics[width=0.077\textwidth]{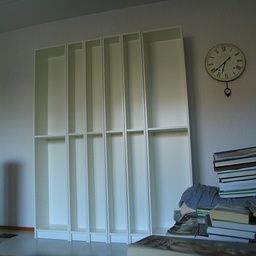} \\
\end{tabular}
\caption{Qualitative results of CGD (ConvNeXt~V2 Base, trained on COCO 2017) on re-encoded COCO test views. Left six: accurate predictions; right six: failure cases.}
\label{fig:qualitative}
\end{figure*}

\begin{figure}[pos=tbp]
\centering
\includegraphics[width=0.9\columnwidth]{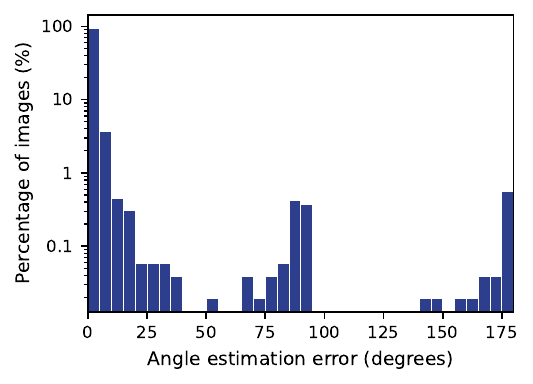}
\caption{Distribution of angle estimation errors of CGD (ConvNeXt~V2 Base, trained on COCO 2017) on re-encoded COCO test views, pooled over five test seeds (log scale). The near-perpendicular (90\textdegree{}~$\pm$~10\textdegree{}) and near-flip ($\geq$170\textdegree{}) error modes visible in the tail together account for 55\% of all failures.}
\label{fig:error_breakdown}
\end{figure}

Figure~\ref{fig:qualitative} shows representative successes and failures of CGD with ConvNeXt~V2 Base trained on COCO 2017 (one of the three runs of Section~\ref{sec:coco2017} and the model we release), on re-encoded test views; the same model underlies Figure~\ref{fig:error_breakdown}, so that the qualitative and quantitative views describe one model. The vast majority of images are corrected to within a few degrees. Among the worst predictions, errors cluster near cardinal rotations (90\textdegree{}, 180\textdegree{}), where the model identifies a plausible orientation axis but selects the wrong polarity or perpendicular direction. We quantify this by pooling per-sample predictions over the five test seeds (5{,}150 predictions, Figure~\ref{fig:error_breakdown}): 97.3\% of predictions err by less than 10\textdegree{}, 0.9\% are near-perpendicular errors (90\textdegree{}~$\pm$~10\textdegree{}), 0.6\% are near-flips ($\geq$170\textdegree{}), and 1.2\% fall elsewhere; the two cardinal error modes thus account for 55\% of all failures, the remainder being intermediate errors. Images whose ground-truth rotation lies near a cardinal orientation fail slightly less often than others (2.0\% vs.\ 2.9\%), indicating that the clustering concerns the error magnitude rather than the ground-truth angle. The cardinal confusions typically occur on images with weak gravitational cues, such as close-up views or symmetric scenes where upright and inverted orientations are visually similar.

\subsection{Ablations and Controls}\label{sec:ablations}

The following experiments each isolate one factor of the protocol or the methods; every paragraph states the dataset it was run on.

\medskip\noindent\textbf{Validation-split sensitivity (DRC-D).}
To verify that the size of the validation split does not drive our conclusions, we re-ran the five base methods on three representative backbones (EfficientViT-B3, MambaOut Base, ConvNeXt V2 Base) with a 20\% validation split instead of the default 10\%, five runs each; Table~\ref{tab:valsplit} lists both settings side by side. The conclusions are unchanged: under the enlarged split CGD is best on all three backbones (1.20\textdegree{}--1.90\textdegree{}) and it stays among the strongest under both, the stable methods mostly shift by a few tenths of a degree (largest individual shift ${\sim}$1\textdegree{}), and direct-angle regression collapses on the same two backbones as before. The intermittent CLS failures also persist under the larger split, though fewer runs collapsed in this experiment (one of ten on the two affected backbones, against four of ten with 10\% validation). Given how few runs each count rests on, we do not read this as an effect of the split size; the CSL variant points to the loss structure as the cause of the failure mode (Section~\ref{sec:analysis}).

\medskip\noindent\textbf{Isolating the representation from the regularization (DRC-D).}
To separate the contribution of the two-dimensional $(\cos\theta, \sin\theta)$ output representation from that of the unit-norm regularization, we also trained the unit vector head without the regularization term on four representative backbones (five runs each; Table~\ref{tab:uvreg}). This naive two-output regression reaches 2.18\textdegree{} (EfficientViT-B3), 2.01\textdegree{} (MambaOut Base), 2.09\textdegree{} (ConvNeXt V2 Base), and 3.37\textdegree{} (ConvNeXt V2 Atto), within 0.7\textdegree{} of the regularized head (2.41\textdegree{}, 1.99\textdegree{}, 1.54\textdegree{}, 2.69\textdegree{}) and nowhere near the collapse of scalar direct-angle regression on the same backbones (7.88\textdegree{}--70.07\textdegree{}). The decisive factor separating UV from DA is therefore the two-dimensional output representation itself; the unit-norm regularization contributes at most marginal, backbone-dependent gains.

\medskip\noindent\textbf{Padding-versus-crop control (DRC-D).}
As a control for geometric cues at the image boundary (Section~\ref{sec:experiments}), we trained the two best configurations of the base methods with rotation-plus-black-padding instead of cropping (five runs each). In the padded setting the image boundary directly encodes the applied angle, so it shows what such a cue contributes when it is available. Accuracy is essentially unchanged for CLS+EfficientViT-B3 (1.23\textdegree{}~$\to$~1.24\textdegree{}) and clearly better for CGD+MambaOut Base (1.24\textdegree{}~$\to$~0.86\textdegree{}, a 31\% reduction), consistent with the trivially visible boundary cue in the padded images. The padded results thus show that an explicit boundary cue can improve accuracy considerably when it is available, as it does for one of the two configurations, whereas the cropped views do not offer this cue. The crop geometry itself is in any case only weakly informative: the inscribed-rectangle dimensions depend on the angle solely through $|\sin\theta|$ and $|\cos\theta|$, so they are invariant under $\theta \to -\theta$ and $\theta \to 180^\circ \pm \theta$. The crop shape is therefore four-fold ambiguous and cannot by itself supply a full-range angle. This control concerns boundary and crop-shape cues only: both of its conditions rotate the decoded source image and therefore carry the rotated block grid of Section~\ref{sec:traces}, which it cannot detect and which is examined there.

\medskip\noindent\textbf{Decoding resolution (DRC-D and COCO).}
Since the classification variants and CGD decode via argmax over 360 bins, their output resolution is 1\textdegree{}, with an expected quantization error of 0.25\textdegree{}. To test whether this limits accuracy, we additionally decode these heads with a soft-argmax (probability-weighted circular mean over bins); Table~\ref{tab:softargmax} in Appendix~\ref{app:ablations} lists both decodings for every evaluation that recorded them. On the COCO grid the effect is small throughout and not uniformly beneficial. CGD changes by $-0.06$\textdegree{} to $+0.01$\textdegree{} (CGD+ConvNeXt~V2 Base 3.52\textdegree{}~$\to$~3.46\textdegree{}), and the three-run means of Section~\ref{sec:coco2017} improve slightly on both datasets (3.88\textdegree{}~$\to$~3.80\textdegree{} and 3.76\textdegree{}~$\to$~3.68\textdegree{}). The CSL cells change by $-0.07$\textdegree{} to $+0.07$\textdegree{}, and the plain cross-entropy cells become slightly worse on all three backbones (by 0.05\textdegree{}--0.20\textdegree{}, the largest change on the grid). The DRC-D evaluations that recorded both decodings (the CSL sweep) show the same picture at higher accuracy: shifts between $-0.10$\textdegree{} and $+0.61$\textdegree{}, with the best configuration (CSL+MambaOut Tiny) moving from 0.81\textdegree{} to 0.75\textdegree{}. Quantization is thus dominated by other error sources at both scales, and binning resolution is not a bottleneck at current accuracy levels; all values reported elsewhere in this paper use argmax decoding.

\subsection{Training Cost}\label{sec:cost}

Table~\ref{tab:cost} reports the measured training cost on a single A100 for all three datasets. On DRC-D a run takes minutes; on COCO the same protocol takes hours, roughly 9 to 20 per run on COCO 2014 depending on the method and 8 to 22 on COCO 2017 depending on the backbone. The COCO timings were measured for the runs of the conventional protocol on a dedicated GPU. The trace-free runs shared the GPU with other jobs, so their wall-clock times are not comparable, and we report their epoch counts instead: averaged over the three backbones, CGD and CSL needed 55 epochs each, DA 72, plain cross-entropy 79, UV 87 and PSC 93. This asymmetry between the regimes motivates the evaluation protocol of Section~\ref{sec:experiments}: five training runs per configuration are affordable on DRC-D, whereas on COCO each grid configuration is trained once (Section~\ref{sec:limitations}). Two consistent observations: the cheapest methods to train are CGD and the CSL variant, with CGD posting the lowest median wall-clock on DRC-D, CSL the fewest epochs there, and both the fewest epochs on COCO under either protocol; and the wide DRC-D ranges reflect heterogeneity across backbones (1.2M--87.7M parameters) rather than run-to-run noise. Circular smooth labels therefore cost no more to train than plain cross-entropy at either scale.

\begin{table}[pos=tbp]
\centering
\caption{Measured training cost on a single A100. DRC-D and COCO 2014 are shown side by side for each method; COCO 2017, for which only CGD was trained, follows below in the same hour units. DRC-D: median [min--max] over all 80 runs per method (16 backbones $\times$ 5 seeds); the wide ranges reflect backbone heterogeneity (1.2M--87.7M parameters) rather than run-to-run noise. COCO 2014: mean over the three grid backbones, one training run each. COCO 2017: one training run per backbone. The COCO values were measured for the runs of the conventional protocol (Section~\ref{sec:traces}); epoch counts of the trace-free runs are given in the text.%The minutes-versus-hours contrast motivates the five-run DRC-D versus single-run COCO evaluation protocol.
}
\label{tab:cost}
\setlength{\tabcolsep}{4pt}
\renewcommand{\arraystretch}{0.95}
\footnotesize
\begin{tabular}{llrrr}
\toprule
& \multicolumn{2}{c}{DRC-D} & \multicolumn{2}{c}{COCO 2014} \\
\costgrouprules
Configuration & Time (min) & Epochs & Time (h) & Epochs \\
\midrule
DA        & 10.2 [0.6--31.0]           & 96 & 12.9 & 90 \\
UV        & \phantom{0}7.6 [3.1--44.3] & 70 & 15.2 & 105 \\
PSC       & \phantom{0}7.7 [2.9--54.2] & 72 & 20.1 & 111 \\
CLS (CE)  & \phantom{0}6.9 [1.7--37.5] & 68 & 12.9 & 81 \\
CLS (CSL) & \phantom{0}6.2 [2.2--23.8] & 55 & 8.6  & 58 \\
CGD       & \phantom{0}6.0 [2.1--21.4] & 57 & 9.8  & 64 \\
\midrule
\multicolumn{5}{l}{\textit{COCO 2017 (hours), CGD only}} \\
\multicolumn{2}{l}{EfficientViT-B3}  & & 8.0  & 47 \\
\multicolumn{2}{l}{ConvNeXt~V2 Base} & & 13.2 & 60 \\
\multicolumn{2}{l}{MambaOut Base}    & & 21.9 & 86 \\
\bottomrule
\end{tabular}
\end{table}

\section{Discussion}\label{sec:analysis}

\medskip\noindent\textbf{Classification and CGD at the top: peak accuracy versus robustness.}
Among the five base methods, classification trained with plain cross-entropy achieves the best single-configuration result on DRC-D (EfficientViT-B3: 1.23\textdegree{}), and its peak performance can be understood from two properties. First, discretizing 360\textdegree{} into bins removes the boundary discontinuity from the regression target: no numerical wrap-around exists in the label space, so the loss never produces the conflicting gradients that plague scalar regression at 0\textdegree{}/360\textdegree{}. We emphasize that this does not remove the circular structure from the learning problem altogether: standard cross-entropy treats the bins as unordered classes, so the adjacency of bins 0 and 359 (and indeed of any neighboring bins) is not encoded in the objective. Circular label-smoothing schemes such as CSL~\cite{yangArbitraryOrientedObjectDetection2020} reintroduce exactly this neighborhood structure. Second, the classification objective takes the same form as the ImageNet pretraining task, so the backbone's learned feature representations are naturally compatible with a classification head, requiring no repurposing of the feature space. However, plain cross-entropy is not uniformly reliable across backbones: several architecture-approach combinations exhibit training failures across multiple runs. A plausible explanation for this instability is the interaction of cross-entropy with the pretrained feature space on a small dataset: cross-entropy aggressively concentrates probability mass on a single class, and when the backbone's pretrained representations do not readily separate rotation-relevant cues, optimization can become trapped in poor regions of the loss landscape before meaningful orientation features emerge, with the missing neighborhood structure between bins offering no partial credit for nearly-correct predictions. Architectures whose inductive biases happen to surface orientation-relevant features early in fine-tuning appear more resistant to this failure mode, which could explain the strong dependence on the specific backbone rather than on model capacity alone. The CSL experiment (CSL column of Table~\ref{tab:overview_all}) corroborates this account experimentally: restoring angular neighborhood structure through circularly smoothed labels eliminated every observed failure case and improved the mean error on 14 of the 16 backbones, which points to the unsmoothed loss, rather than the backbone, as the operative cause. Smoothed labels differ from one-hot labels in more than their neighborhood structure, however, so this experiment does not isolate that structure from the other effects of softer targets. The validation-split sensitivity experiment (Section~\ref{sec:ablations}) adds a complementary observation: with a larger validation split the failures occur less frequently but do not disappear, consistent with a stochastic optimization failure, which CSL removes. Figure~\ref{fig:loss_curves} shows the failure mode at the level of individual training runs. We note that the failures are genuinely stochastic rather than seed-determined: when re-running historically failing seeds for that diagnostic, most re-runs converged, and the failing example was obtained only after several attempts.

CGD is nearly tied at the top on DRC-D (MambaOut Base: 1.24\textdegree{}) while remaining reliably stable across all tested backbones, arguably making it the more practical choice for deployment among the base methods. This robustness admits the same explanation: CGD's circular-Gaussian targets already encode angular neighborhood structure, awarding partial credit to near-miss bins just as CSL's smoothed labels do, so the ingredient whose absence destabilizes plain cross-entropy is built into its supervision. The classification-family results thus reduce to one principle: the heads whose supervision encodes angular proximity (CSL and CGD) converged in every run, while the only head without it failed intermittently.

PSC provides a continuous code and reversible decoding, which is attractive when the code must be blended with other parameters (as in oriented detection)~\cite{yuPhaseShiftingCoderPredicting2023,yuBoundaryDiscontinuityAngle2024,xuRethinkingBoundaryDiscontinuity2024}. On global orientation, CGD tends to have an edge (on DRC-D and on two of the three backbones at scale), plausibly reflecting its smoothed distributional supervision, with its distributional output as a further practical benefit; PSC remains competitive on larger backbones.
Even with circular losses, the direct scalar formulation is brittle: gradients conflict when predictions straddle the boundary, and the network can settle into large-error modes. Our results quantify this effect across many backbones.

\medskip\noindent\textbf{Why the CSL variant dominates DRC-D but only matches the other classification heads on COCO.}
CSL is the best configuration on 8 of 16 DRC-D backbones and sets the study's best small-data result (0.81\textdegree{}), yet on COCO it is one of three heads of the classification family with comparable performance: it is best on one backbone, CGD and plain cross-entropy on one each (Table~\ref{tab:coco_all}). Three factors, all visible in our data, contribute to the difference. First, the two tables treat failed runs differently: the DRC-D means include the collapsed runs of plain cross-entropy, whereas on COCO a visibly failed run is repeated and the successful run reported (Section~\ref{sec:limitations}). The collapse that CSL prevents therefore weighs on the plain cross-entropy column of Table~\ref{tab:overview_all} but not on that of Table~\ref{tab:coco_all}, although it occurred at scale as well: one of the four plain cross-entropy runs of the COCO grid remained at chance level (Section~\ref{sec:coco_matrix}), while CSL and CGD converged in every COCO run. Second, the two regimes differ in protocol: the DRC-D models were trained and tested under the conventional protocol, in which the binned heads, CSL most of all, draw part of their fine precision from the block grid (Table~\ref{tab:drcd_reenc}); on re-encoded DRC-D views CSL remains the strongest head by median, but its lead over CGD shrinks from 0.52\textdegree{} to 0.18\textdegree{}. Third, the error profiles decompose differently at the two scales. On COCO, CSL is the more precise head on typical images, with per-cell median errors (0.74\textdegree{}--1.33\textdegree{}) below CGD's (1.30\textdegree{}--1.68\textdegree{}) on every backbone, and the two are close up to the 95th percentile (7.0\textdegree{}--8.6\textdegree{} versus 6.4\textdegree{}--8.8\textdegree{}); these per-cell statistics are computed from the per-seed evaluations, whereas Table~\ref{tab:coco_all} tabulates them only for the per-backbone best method. The mean error on COCO is dominated by the rare errors near 90\textdegree{} and 180\textdegree{} on ambiguous images, which CSL's precision does not reduce: its RMSE (16.9\textdegree{}--19.6\textdegree{}) is no lower than CGD's (14.1\textdegree{}--19.7\textdegree{}). On DRC-D, whose curated scenes with clear upright cues make such failures rare, precision on typical images dominates instead. These explanations come with the caveats that the smoothing width is held at the value adopted from the original work on both datasets, that the COCO cells are single training runs (Section~\ref{sec:limitations}), and that two of the most precise classification cells miss one criterion of the content-free probe (Section~\ref{sec:traces}); per-dataset tuning of the smoothing might shift this balance.

\medskip\noindent\textbf{Architecture-approach compatibility.}
Not all combinations are equally effective. Within a family, the larger backbone is usually but not always the better one in Table~\ref{tab:overview_all}: for the four stable methods (UV, PSC, CSL and CGD) it has the lower error in 26 of the 32 pairs. Across families the picture is less regular, and the interaction between method and architecture matters: a well-matched smaller backbone can outperform a larger one of another family (CSL reaches 0.81\textdegree{} with MambaOut Tiny and 2.16\textdegree{} with ViT-Base), which is why the comparison is run across architectures.

\section{Practical Guidance}\label{sec:guidance}
One result holds across both regimes: classification heads outperform regression heads, taking the lowest error on thirteen of the sixteen small-data backbones and, on re-encoded test views, on all three tested at scale, most reliably when the objective encodes angular neighborhood structure. Beyond that, the guidance differs by data regime. In the small-data regime, classification with circular smooth labels is the first choice: CSL with MambaOut Tiny achieves the best result in this study (0.81\textdegree{}), and CSL converged in every run on every tested backbone. Plain cross-entropy classification can match this tier on well-matched backbones (EfficientViT-B3: 1.23\textdegree{}) but carries a stochastic instability risk that CSL removes at negligible cost. At scale, the three classification heads perform comparably, and we recommend CGD or CSL over plain cross-entropy, which again failed to train in one run: neither CGD nor CSL failed to converge in any of our experiments, and CGD with ConvNeXt~V2 Base is the best configuration we trained (3.76\textdegree{} on COCO 2017). Whatever the head, training views produced by rotating stored photographs should be re-encoded after rotation and cropping, and a trained model should be checked with a content-free probe before its accuracy is trusted (Section~\ref{sec:traces}). A model trained this way is best applied to images in the form in which they are stored; inputs that were never re-encoded lie slightly outside its training distribution. CGD is also a natural candidate when a distributional output is wanted (e.g., for rejecting uncertain predictions), although we did not evaluate the calibration of its predicted probabilities. PSC may be preferred when angles must be fused with other continuous parameters within a single code, as in oriented object detection. Given that training outcomes can vary across runs, reporting means over multiple runs, as we do on DRC-D, provides more reliable performance estimates for final deployment decisions.

\section{Limitations and Future Work}\label{sec:limitations}

Several limitations should be acknowledged when interpreting our results. Deep learning training is inherently non-deterministic due to random weight initialization, data shuffling, and hardware-specific floating-point operations. To quantify this variability, all results in Table~\ref{tab:overview_all} report means and standard deviations across five independent training runs with different random seeds. This multi-run protocol substantially improves reliability compared to single-run evaluations: for some architecture-approach combinations, individual runs can vary by several degrees in MAE, while five-run means are considerably more stable. Standard deviations in the table reflect the true run-to-run variability and should be consulted alongside means when interpreting results.

This protection does not extend to our COCO results, and the distinction matters when reading them. Because a single COCO training run takes roughly 9 to 20 hours depending on the method (Table~\ref{tab:cost}), repeating every configuration five times was not feasible within a reasonable time frame; each cell of the COCO grid is therefore a single training run, and the standard deviations in the grid block of Table~\ref{tab:coco_all} describe variability across test rotations only, not how much a second training run of the same configuration would differ. Where a training run visibly failed, collapsing to near-random accuracy or stalling far above the level its configuration reaches elsewhere, we repeated it and report the successful run; this concerned one run, plain cross-entropy on ConvNeXt~V2 Base (Section~\ref{sec:coco_matrix}). This applies only where the configuration converges elsewhere; the direct-angle cells that fail on two backbones are reported as they are, because that is the behaviour of the configuration rather than of a single run. How much a second run can differ is shown by the configurations we trained more than once: the three runs of CGD with ConvNeXt~V2 Base span 0.72\textdegree{} on COCO 2014 and 0.11\textdegree{} on COCO 2017 (Section~\ref{sec:coco2017}), and for six further grid configurations a run of a preliminary round under the same protocol exists, whose checkpoints were not retained; it differs from the reported run by 0.08\textdegree{} to 1.56\textdegree{}. The grid reports the runs of the complete round throughout. The COCO numbers should accordingly be read as indicating tendencies rather than a precise ranking: the broad structure is dependable (direct regression fails, the structured methods cluster between roughly 3.5\textdegree{} and 5.3\textdegree{}), and the family ordering found on DRC-D recurs on all three backbones on re-encoded views, though by margins of 0.6\textdegree{} to 1.3\textdegree{} that are of the size of the differences between two runs of one configuration, whereas fine distinctions between neighbouring cells, including which classification head is best on a given backbone, rest on one training run each. A related practical note: an early-stopping criterion that counts any improvement in validation MAE, however small, can keep long COCO runs oscillating within noise instead of terminating, which is why the COCO runs require an improvement of at least 0.1\textdegree{} (Section~\ref{sec:experiments}).

Our most extensive comparison is the small-data regime, DRC-D, which contains only 1{,}474 training images. While this small size is what makes the 96-configuration, five-seed study computationally feasible, it also means that method rankings established there may shift on larger and more diverse datasets. The COCO 2014 grid tests every method at scale and finds the broad pattern preserved, but it does so on three representative backbones; extending the full sixteen-backbone grid to a large dataset remains open, and the backbone-level conclusions at scale accordingly rest on a narrower base than those on DRC-D.

The processing-trace analysis of Section~\ref{sec:traces} has limits of its own. Our checks target the two traces we could identify, the JPEG block grid and the interpolation pattern of the rotation, and a model that passes them is not thereby shown to be free of every low-level cue. Two grid cells miss the share criterion of the content-free probe, one of them clearly, and each COCO 2017 run misses it in one condition by at most 1.5 points (Appendix~\ref{app:traces}). The two cells, plain cross-entropy on MambaOut Base and CSL on ConvNeXt~V2 Base, are also the most precise on re-encoded views (median errors of 0.33\textdegree{} and 0.74\textdegree{}), and although the photograph controls attribute none of that precision to the rotated source grid or to the interpolation, we could not identify what the probe responds to in these models. The trace-free protocol changes two things at once, the re-encoding and the interpolation kernel, and we did not ablate them separately. The re-encoded test condition simulates storage with a standard codec; it is not a test on photographs taken with a physically rotated camera. For DRC-D, only the best run of each configuration was retained, so the re-encoded values quoted in Section~\ref{sec:traces} are single-run values rather than five-run means, and the trace-free protocol itself was applied on COCO only: Table~\ref{tab:overview_all} reports models trained and tested under the conventional protocol.

The two families also differ in the architecture of their heads: the regression methods use a small MLP and the classification methods a linear layer (Section~\ref{sec:experiments}), so the family comparison is one between complete heads and not between output representations alone. Hyperparameter optimization represents another limitation. While we identified suitable learning rates for different architectures, optimal hyperparameters depend on complex interactions between architecture design, optimizer choice, dataset characteristics, and batch size. The learning rates and other training parameters used may not represent the optimal configuration for each architecture-approach combination, potentially affecting the relative comparisons presented here. Relatedly, probabilistic approaches like CGD introduce additional hyperparameters, most notably the Gaussian width $\sigma$ and the number of bins, whose defaults worked well across architectures in our experiments but were not exhaustively tuned. The same applies to the circular smoothing width used by CSL, which we keep at the value adopted from the original work on both datasets.

Finally, our study focuses exclusively on global in-plane rotation estimation of natural images with clear upright indicators; scenes with inherently ambiguous orientation (aerial views, abstract textures, strong symmetries) fall outside the scope of the evaluated datasets, and the qualitative failure cases in Figure~\ref{fig:qualitative} suggest that weak gravitational cues remain the dominant practical failure mode.

Several directions follow for future work. Extending the multi-run protocol from the best configuration to every COCO cell is the most useful reliability improvement. Beyond that, extending to full 3D pose estimation (yaw/pitch), incorporating semantic cues such as horizon lines, adaptive binning strategies and learned distribution widths for CGD, per-dataset tuning of the CSL smoothing width, and bridging to content-preserving warps without explicit angle prediction~\cite{nieDeepRotationCorrection2023} represent promising directions.

\section{Conclusion}\label{sec:conclusion}

Modeling circular structure is essential for reliable angle prediction. In a comparison of five circular-aware methods across sixteen architectures, classification approaches outperform regression approaches in both regimes, attaining the lowest error on thirteen of the sixteen DRC-D backbones and, on re-encoded test views, on all three backbones tested at scale; the wins concentrate in the heads whose supervision encodes angular neighborhood structure. On DRC-D the best result comes from classification with circular smooth labels (MambaOut Tiny: 0.81\textdegree{}), nominally ahead of plain cross-entropy classification (EfficientViT-B3: 1.23\textdegree{}) and CGD (MambaOut Base: 1.24\textdegree{}), which are statistically indistinguishable from one another; of those two, CGD is the more robust across backbone choices. The margin matters less than the mechanism: the classification instabilities observed on several backbones are linked to the missing circular neighborhood structure in plain cross-entropy, and restoring it through smoothed labels eliminates every observed failure; CGD, which like CSL never failed to train in any of our experiments, carries the same structure in its Gaussian targets. At scale, the comparison first required repairing the protocol: rotating stored photographs leaves the rotated JPEG block grid as a cue to the applied angle, which the binned classification heads learned on COCO and the regression heads largely did not, and which a single re-encoding step after rotation suppresses. Under the resulting trace-free protocol, across three representative backbones on COCO 2014, the same family ordering recurs with smaller margins and with comparable performance of the three classification heads. The best configuration, CGD with ConvNeXt~V2 Base, reaches 3.88\textdegree{} MAE on COCO 2014 and 3.76\textdegree{} on the larger COCO 2017 set (means of three training runs); the best method--backbone pairing thus differs between the two regimes, so backbone selection does not transfer together with the family ordering.
These findings, together with our reproducible framework, contribute to future work in circular regression, oriented object detection, and pose estimation.

\printcredits

\section*{Acknowledgements}
The author thanks Vladimir Ilyash (ORTUS AI) for reporting the sensitivity of the released model to JPEG re-encoding and for sharing a reproduction script.

\section*{Declaration of generative AI and AI-assisted technologies in the manuscript preparation process}
During the preparation of this work the author used Claude and Claude Code (Anthropic) in order to assist with method implementation, experiment tooling, data analysis, and the drafting and revision of text. After using these tools, the author reviewed and edited the content as needed and takes full responsibility for the content of the published article.

\section*{Declaration of competing interest}
The author is the owner of MAXI solutions e.U., a software company whose work includes computer-vision applications. The author declares no other competing financial interests or personal relationships that could have appeared to influence the work reported in this paper.

\section*{Funding}
This research did not receive any specific grant from funding agencies in the public, commercial, or not-for-profit sectors.

\section*{Data and Code Availability}
Code and data are available at \url{https://github.com/maxwoe/image-rotation-angle-estimation}, including the trace-free training option and the scripts of the processing-trace checks. The trained model is available at \url{https://huggingface.co/maxwoe/image-rotation-angle-estimation}, and a demo at \url{https://huggingface.co/spaces/maxwoe/image-rotation-angle-estimation}.

\appendix

\section{Ablation Details}\label{app:ablations}

This appendix collects the detailed result tables for the ablations and controls of Section~\ref{sec:ablations}.

\medskip\noindent\textbf{Validation split.}
Table~\ref{tab:valsplit} compares the 10\% and 20\% validation-split settings of the sensitivity experiment.

\medskip\noindent\textbf{Unit-vector regularization.}
Table~\ref{tab:uvreg} details the unit-vector regularization ablation.

\medskip\noindent\textbf{Decoding resolution.}
Table~\ref{tab:softargmax} lists both decodings for every evaluation that recorded them: the DRC-D CSL sweep and all binned heads of the COCO grids (the regression heads DA, UV, and PSC produce continuous outputs and involve no decoding choice). The original DRC-D grid predates the soft-argmax implementation, so CLS (CE) and CGD DRC-D cells are not available in both decodings. All values reported in the main text use argmax.

\begin{table}[pos=htbp]
\centering
\caption{Validation-split sensitivity on DRC-D: MAE\textdegree{} as mean (standard deviation) over five training runs per cell, under the default 10\% and the enlarged 20\% validation split, for the five base methods (the CSL variant was evaluated subsequently and is not part of this experiment). The 10\% values are the corresponding cells of Table~\ref{tab:overview_all}. Bracketed counts give the number of runs, out of five, whose MAE exceeds 10\textdegree{}; those cells are bimodal, so their means should be read together with that count rather than on their own.}
\label{tab:valsplit}
\setlength{\tabcolsep}{4pt}
\renewcommand{\arraystretch}{0.95}
\footnotesize
\begin{tabular}{llcc}
\toprule
Backbone & Method & 10\% val & 20\% val \\
\midrule
\multicolumn{4}{l}{\textit{ConvNeXt V2 Base}} \\
& DA & 61.42(8.16) [5/5] & 66.64(9.76) [5/5] \\
& UV & 1.54(0.36) & 2.40(1.10) \\
& PSC & 1.47(0.16) & 1.84(0.18) \\
& CLS (CE) & 20.71(33.63) [1/5] & 16.85(31.41) [1/5] \\
& CGD & 2.22(1.23) & 1.20(0.12) \\
\midrule
\multicolumn{4}{l}{\textit{EfficientViT-B3}} \\
& DA & 8.25(1.84) [2/5] & 7.75(1.59) [0/5] \\
& UV & 2.41(0.35) & 2.29(0.18) \\
& PSC & 2.74(1.53) & 2.73(0.81) \\
& CLS (CE) & 1.23(0.61) & 2.15(0.35) \\
& CGD & 1.93(0.33) & 1.90(0.35) \\
\midrule
\multicolumn{4}{l}{\textit{MambaOut Base}} \\
& DA & 70.07(8.65) [5/5] & 39.20(16.73) [5/5] \\
& UV & 1.99(0.38) & 2.23(0.65) \\
& PSC & 2.57(1.21) & 2.91(0.55) \\
& CLS (CE) & 55.15(42.98) [3/5] & 1.93(1.18) [0/5] \\
& CGD & 1.24(0.53) & 1.52(0.29) \\
\bottomrule
\end{tabular}
\end{table}

\begin{table}[pos=htbp]
\centering
\caption{Unit-vector regularization ablation on DRC-D: MAE\textdegree{} as mean (standard deviation) over five training runs, for the UV head with and without the unit-norm regularization term. The regularized values are the UV cells of Table~\ref{tab:overview_all}.}
\label{tab:uvreg}
\setlength{\tabcolsep}{5pt}
\renewcommand{\arraystretch}{0.95}
\footnotesize
\begin{tabular}{lcc}
\toprule
Backbone & With regularization & Without \\
\midrule
ConvNeXt V2 Atto & 2.69(0.41) & 3.37(0.76) \\
ConvNeXt V2 Base & 1.54(0.36) & 2.09(0.91) \\
EfficientViT-B3 & 2.41(0.35) & 2.18(0.44) \\
MambaOut Base & 1.99(0.38) & 2.01(0.38) \\
\bottomrule
\end{tabular}
\end{table}

\begin{table}[pos=htbp]
\centering
\caption{MAE under argmax and soft-argmax decoding for every evaluation that recorded both. DRC-D values are means over five training runs; COCO values are means over five test seeds on re-encoded test views, from a single training run per grid cell and averaged over three training runs in the last two rows. $\Delta$ = soft-argmax minus argmax (negative = soft-argmax better).}
\label{tab:softargmax}
\setlength{\tabcolsep}{4pt}
\renewcommand{\arraystretch}{0.92}
\footnotesize
\resizebox{\columnwidth}{!}{%
\begin{tabular}{llccc}
\toprule
Backbone & Method & Argmax & Soft-argmax & $\Delta$ \\
\midrule
\multicolumn{5}{l}{\textit{DRC-D (CSL sweep)}} \\
ConvNeXt V2 Atto & CLS (CSL) & 3.92\textdegree{} & 4.26\textdegree{} & $+$0.34\textdegree{} \\
ConvNeXt V2 Base & CLS (CSL) & 0.93\textdegree{} & 0.83\textdegree{} & $-$0.10\textdegree{} \\
EdgeNeXt XX-Small & CLS (CSL) & 2.33\textdegree{} & 2.52\textdegree{} & $+$0.19\textdegree{} \\
EdgeNeXt Base & CLS (CSL) & 1.65\textdegree{} & 1.55\textdegree{} & $-$0.09\textdegree{} \\
EfficientNetV2-RW~T & CLS (CSL) & 3.37\textdegree{} & 3.32\textdegree{} & $-$0.05\textdegree{} \\
EfficientNetV2-RW~M & CLS (CSL) & 1.90\textdegree{} & 2.19\textdegree{} & $+$0.29\textdegree{} \\
EfficientViT-B0 & CLS (CSL) & 6.20\textdegree{} & 6.46\textdegree{} & $+$0.26\textdegree{} \\
EfficientViT-B3 & CLS (CSL) & 1.59\textdegree{} & 2.20\textdegree{} & $+$0.61\textdegree{} \\
FocalNet Tiny LRF & CLS (CSL) & 1.82\textdegree{} & 1.78\textdegree{} & $-$0.05\textdegree{} \\
FocalNet Base LRF & CLS (CSL) & 2.23\textdegree{} & 2.21\textdegree{} & $-$0.03\textdegree{} \\
MambaOut Tiny & CLS (CSL) & 0.81\textdegree{} & 0.75\textdegree{} & $-$0.06\textdegree{} \\
MambaOut Base & CLS (CSL) & 1.05\textdegree{} & 1.05\textdegree{} & $-$0.00\textdegree{} \\
Swin Tiny & CLS (CSL) & 1.48\textdegree{} & 1.63\textdegree{} & $+$0.15\textdegree{} \\
Swin Base & CLS (CSL) & 1.02\textdegree{} & 0.92\textdegree{} & $-$0.10\textdegree{} \\
ViT-Tiny & CLS (CSL) & 6.76\textdegree{} & 6.92\textdegree{} & $+$0.17\textdegree{} \\
ViT-Base & CLS (CSL) & 2.16\textdegree{} & 2.06\textdegree{} & $-$0.10\textdegree{} \\
\midrule
\multicolumn{5}{l}{\textit{COCO 2014}} \\
ConvNeXt V2 Base & CLS (CE) & 4.91\textdegree{} & 5.05\textdegree{} & $+$0.14\textdegree{} \\
ConvNeXt V2 Base & CLS (CSL) & 3.60\textdegree{} & 3.67\textdegree{} & $+$0.07\textdegree{} \\
ConvNeXt V2 Base & CGD & 3.52\textdegree{} & 3.46\textdegree{} & $-$0.06\textdegree{} \\
EfficientViT-B3 & CLS (CE) & 4.51\textdegree{} & 4.71\textdegree{} & $+$0.20\textdegree{} \\
EfficientViT-B3 & CLS (CSL) & 4.74\textdegree{} & 4.75\textdegree{} & $+$0.02\textdegree{} \\
EfficientViT-B3 & CGD & 4.75\textdegree{} & 4.71\textdegree{} & $-$0.03\textdegree{} \\
MambaOut Base & CLS (CE) & 4.58\textdegree{} & 4.63\textdegree{} & $+$0.05\textdegree{} \\
MambaOut Base & CLS (CSL) & 4.09\textdegree{} & 4.02\textdegree{} & $-$0.07\textdegree{} \\
MambaOut Base & CGD & 4.80\textdegree{} & 4.81\textdegree{} & $+$0.01\textdegree{} \\
\midrule
\multicolumn{5}{l}{\textit{CGD with ConvNeXt V2 Base, means of three training runs}} \\
COCO 2014 & CGD & 3.88\textdegree{} & 3.80\textdegree{} & $-$0.08\textdegree{} \\
COCO 2017 & CGD & 3.76\textdegree{} & 3.68\textdegree{} & $-$0.09\textdegree{} \\
\bottomrule
\end{tabular}}
\end{table}

\section{Processing-Trace Checks and Results Without Re-Encoding}\label{app:traces}

This appendix specifies the checks of Section~\ref{sec:traces} and their outcomes, and lists the trace-free models on test views without re-encoding (Table~\ref{tab:coco_noreenc}). The pass criteria of the probe and of the two photograph controls were fixed in the evaluation scripts before the respective evaluations were run; the cue-conflict test reports a share and has no criterion.

\medskip\noindent\textbf{Content-free probe.}
One thousand isotropic noise textures of $640\times480$ pixels are rotated and cropped by the pipeline, either as generated or after being saved once as JPEG at quality 90, 75 or 50, which stamps a block grid that then rotates with the texture. Two statistics are computed from the predictions $\hat{\theta}$ and the applied angles $\theta$: the share of predictions within 5\textdegree{} of $\theta$ modulo 90\textdegree{} (11.1\% by chance), and $\rho_4$, the mean of $\cos 4(\hat{\theta}-\theta)$, which is 0 for a model that ignores the grid and 1 for a model locked onto it. A model passes a condition if the upper bound of the 95\% bootstrap interval of $\rho_4$ is below 0.20, the limit used by ORTUS AI~\cite{ortusaiRightWayUp2026}, and that of the share exceeds chance by less than 5 points. A second set of conditions mirrors the re-encoded test views: textures are stamped at quality 95 or 75, or not at all, rotated and cropped, and then re-encoded at quality 90. For the DRC-D models of Section~\ref{sec:traces} the probe uses 400 textures of $512\times384$ pixels, and the shares are reported as measured, without the pass criteria.

\medskip\noindent\textbf{Cue conflict.}
On 300 test photographs, the source grid is suppressed by a Gaussian blur ($\sigma = 2.5$ pixels) and a grid of quality 75 is stamped at an offset of $-20$\textdegree{}, $+20$\textdegree{} or $+45$\textdegree{} from the content before the final rotation. We report the share of predictions that are closer, modulo 90\textdegree{}, to the angle indicated by the grid than to that of the content.

\medskip\noindent\textbf{Source-grid reliance and rotation residue on photographs.}
Both controls use all test photographs and the five test seeds, end with one JPEG save at quality 90, and compare Acc@2\textdegree{} between two renderings of the same views, with a bootstrap over images; a model passes if the upper bound of the 95\% interval of the difference is below 5 points. For the source grid, the photograph is blurred before rotation ($\sigma = 2.5$), which removes its grid, and in the second rendering additionally re-encoded at quality 95 before rotation, which puts a rotated grid back at identical detail. For the rotation residue, the normal bilinear rotation is compared with a rotation performed at four-fold resolution and averaged down, which leaves no interpolation pattern at the pixel scale.

\medskip\noindent\textbf{Outcomes.}
All 18 grid cells of Table~\ref{tab:coco_all} and the six runs of the best configuration pass the $\rho_4$ criterion in every probe condition, and all pass both photograph controls, with differences between $-1.3$ and $+0.6$ points for the source grid and between $-2.1$ and $+2.3$ points for the rotation residue (upper bounds of at most 3.3). The two conventional-protocol models we subjected to the source-grid control, CGD and plain cross-entropy with ConvNeXt~V2 Base, gain 23.2 and 28.6 points when the final save is lossless, as in their own protocol. The share criterion of the probe, a limit of 5 points, is missed in the following cases: CSL with ConvNeXt~V2 Base (upper bounds of 5.8 to 8.4 points on quality-90 textures and on the conditions that mirror the test views), plain cross-entropy with MambaOut Base (10.0 to 14.2 points on the conditions that mirror the test views), and each of the three COCO 2017 runs of CGD with ConvNeXt~V2 Base in one condition (5.1, 5.2 and 6.5 points); the three COCO 2014 runs pass every condition. Under cue conflict, the offset grid attracts at most 6\% of the predictions of every converged model with a structured head.

\medskip\noindent\textbf{DRC-D.}
Table~\ref{tab:drcd_reenc} gives, by head, the outcome of the check of the DRC-D models on re-encoded views (Section~\ref{sec:traces}). For the five structured heads, Acc@10\textdegree{} changes by at most 0.9 points and the median error by at most 0.39\textdegree{} in any configuration. On textures without a grid and with a quality-90 grid, every head places 10.8\% to 12.2\% of its predictions within 5\textdegree{} of the grid angle on average, against 11.1\% by chance.

\begin{table}[pos=htbp]
\centering
\caption{The trace-free models of Table~\ref{tab:coco_all} on test views without re-encoding, the test condition of the conventional protocol: MAE\textdegree{} as mean (standard deviation) across five test seeds for the grid, and across three training runs for the last two rows. These views lie slightly outside the training distribution of the trace-free models. OAD-360 reaches 6.10(0.67)\textdegree{} on the same views.}
\label{tab:coco_noreenc}
\setlength{\tabcolsep}{3pt}
\renewcommand{\arraystretch}{0.95}
\footnotesize
\resizebox{\columnwidth}{!}{%
\begin{tabular}{lcccccc}
\toprule
Backbone & DA & UV & PSC & CLS (CE) & CLS (CSL) & CGD \\
\midrule
\multicolumn{7}{l}{\textit{COCO 2014, one training run per cell}} \\
ConvNeXt V2 Base & 64.70(0.56) & 5.13(0.34) & 5.02(0.32) & 6.00(0.26) & 4.86(0.34) & \textbf{4.33(0.28)} \\
EfficientViT-B3 & 11.86(0.44) & 5.76(0.30) & 5.60(0.25) & 5.18(0.34) & \textbf{5.14(0.28)} & 5.58(0.23) \\
MambaOut Base & 67.85(0.76) & 5.32(0.46) & \textbf{4.90(0.10)} & 7.18(0.43) & 4.98(0.32) & 5.41(0.46) \\
\midrule
\multicolumn{7}{l}{\textit{CGD with ConvNeXt V2 Base, three training runs per dataset}} \\
COCO 2014 & -- & -- & -- & -- & -- & 4.59(0.28) \\
COCO 2017 & -- & -- & -- & -- & -- & 4.45(0.05) \\
\bottomrule
\end{tabular}}
\end{table}

\bibliographystyle{elsarticle-num}

\bibliography{main}

\begin{thebibliography}{10}
\expandafter\ifx\csname url\endcsname\relax
  \def\url#1{\texttt{#1}}\fi
\expandafter\ifx\csname urlprefix\endcsname\relax\def\urlprefix{URL }\fi
\expandafter\ifx\csname href\endcsname\relax
  \def\href#1#2{#2} \def\path#1{#1}\fi

\bibitem{zhouContinuityRotationRepresentations2019}
Y.~Zhou, C.~Barnes, J.~Lu, J.~Yang, H.~Li, On the {{Continuity}} of {{Rotation
  Representations}} in {{Neural Networks}}, in: 2019 {{IEEE}}/{{CVF
  Conference}} on {{Computer Vision}} and {{Pattern Recognition}} ({{CVPR}}),
  IEEE, Long Beach, CA, USA, 2019, pp. 5738--5746.
\newblock \href {https://doi.org/10.1109/CVPR.2019.00589}
  {\path{doi:10.1109/CVPR.2019.00589}}.

\bibitem{levinsonAnalysisSVDDeep}
J.~Levinson, C.~Esteves, K.~Chen, N.~Snavely, A.~Kanazawa, A.~Rostamizadeh,
  A.~Makadia, An {{Analysis}} of {{SVD}} for {{Deep Rotation Estimation}}
  (2020).

\bibitem{fischerImageOrientationEstimation2015}
P.~Fischer, A.~Dosovitskiy, T.~Brox, Image {{Orientation Estimation}} with
  {{Convolutional Networks}}, in: J.~Gall, P.~Gehler, B.~Leibe (Eds.), Pattern
  {{Recognition}}, Vol. 9358, Springer International Publishing, Cham, 2015,
  pp. 368--378.
\newblock \href {https://doi.org/10.1007/978-3-319-24947-6_30}
  {\path{doi:10.1007/978-3-319-24947-6_30}}.

\bibitem{majiDeepImageOrientation2020}
S.~Maji, S.~Bose, Deep {{Image Orientation Angle Detection}} (Jun. 2020).
\newblock \href {http://arxiv.org/abs/2007.06709} {\path{arXiv:2007.06709}},
  \href {https://doi.org/10.48550/arXiv.2007.06709}
  {\path{doi:10.48550/arXiv.2007.06709}}.

\bibitem{amjoudTransferLearningAutomatic2022}
A.~B. Amjoud, M.~Amrouch, Transfer {{Learning}} for {{Automatic Image
  Orientation Detection Using Deep Learning}} and {{Logistic Regression}}, IEEE
  Access 10 (2022) 128543--128553.
\newblock \href {https://doi.org/10.1109/ACCESS.2022.3225455}
  {\path{doi:10.1109/ACCESS.2022.3225455}}.

\bibitem{joshiAutomaticPhotoOrientation2017}
U.~Joshi, M.~Guerzhoy, Automatic {{Photo Orientation Detection}} with
  {{Convolutional Neural Networks}}, in: 2017 14th {{Conference}} on
  {{Computer}} and {{Robot Vision}} ({{CRV}}), IEEE, Edmonton, AB, 2017, pp.
  103--108.
\newblock \href {https://doi.org/10.1109/CRV.2017.59}
  {\path{doi:10.1109/CRV.2017.59}}.

\bibitem{dongBlindDetectionCircular2022}
W.~Dong, J.~Zhang, Y.~Zhou, L.~Gao, X.~Zhang, Blind detection of circular image
  rotation angle based on ensemble transfer regression and fused {{HOG}},
  Front. Neurorobot. 16 (2022) 1037381.
\newblock \href {https://doi.org/10.3389/fnbot.2022.1037381}
  {\path{doi:10.3389/fnbot.2022.1037381}}.

\bibitem{xiaDOTALargeScaleDataset2018}
G.-S. Xia, X.~Bai, J.~Ding, Z.~Zhu, S.~Belongie, J.~Luo, M.~Datcu, M.~Pelillo,
  L.~Zhang, {{DOTA}}: {{A Large-Scale Dataset}} for {{Object Detection}} in
  {{Aerial Images}}, in: 2018 {{IEEE}}/{{CVF Conference}} on {{Computer
  Vision}} and {{Pattern Recognition}}, IEEE, Salt Lake City, UT, 2018, pp.
  3974--3983.
\newblock \href {https://doi.org/10.1109/CVPR.2018.00418}
  {\path{doi:10.1109/CVPR.2018.00418}}.

\bibitem{dingLearningRoITransformer2019}
J.~Ding, N.~Xue, Y.~Long, G.-S. Xia, Q.~Lu, Learning {{RoI Transformer}} for
  {{Oriented Object Detection}} in {{Aerial Images}}, in: 2019 {{IEEE}}/{{CVF
  Conference}} on {{Computer Vision}} and {{Pattern Recognition}} ({{CVPR}}),
  IEEE, Long Beach, CA, USA, 2019, pp. 2844--2853.
\newblock \href {https://doi.org/10.1109/CVPR.2019.00296}
  {\path{doi:10.1109/CVPR.2019.00296}}.

\bibitem{chengOrientedCenterNetRotated2023}
M.~Cheng, A.~Li, D.~Liu, D.~Yao, X.~Liu, Oriented {{CenterNet}}: {{Rotated
  Object Detection}} in {{Remote Sensing Images}}, in: 2023 {{IEEE
  International Conference}} on {{Systems}}, {{Man}}, and {{Cybernetics}}
  ({{SMC}}), IEEE, Honolulu, Oahu, HI, USA, 2023, pp. 2164--2170.
\newblock \href {https://doi.org/10.1109/SMC53992.2023.10394459}
  {\path{doi:10.1109/SMC53992.2023.10394459}}.

\bibitem{pavlloQuaterNetQuaternionbasedRecurrent2018}
D.~Pavllo, D.~Grangier, M.~Auli, {{QuaterNet}}: {{A Quaternion}}-based
  {{Recurrent Model}} for {{Human Motion}} (Jul. 2018).
\newblock \href {http://arxiv.org/abs/1805.06485} {\path{arXiv:1805.06485}},
  \href {https://doi.org/10.48550/arXiv.1805.06485}
  {\path{doi:10.48550/arXiv.1805.06485}}.

\bibitem{yangArbitraryOrientedObjectDetection2020}
X.~Yang, J.~Yan, Arbitrary-{{Oriented Object Detection}} with {{Circular Smooth
  Label}}, in: A.~Vedaldi, H.~Bischof, T.~Brox, J.-M. Frahm (Eds.), Computer
  {{Vision}} -- {{ECCV}} 2020, Vol. 12353, Springer International Publishing,
  Cham, 2020, pp. 677--694.
\newblock \href {https://doi.org/10.1007/978-3-030-58598-3_40}
  {\path{doi:10.1007/978-3-030-58598-3_40}}.

\bibitem{yangDenseLabelEncoding2021}
X.~Yang, L.~Hou, Y.~Zhou, W.~Wang, J.~Yan, Dense {{Label Encoding}} for
  {{Boundary Discontinuity Free Rotation Detection}}, in: 2021 {{IEEE}}/{{CVF
  Conference}} on {{Computer Vision}} and {{Pattern Recognition}} ({{CVPR}}),
  IEEE, Nashville, TN, USA, 2021, pp. 15814--15824.
\newblock \href {https://doi.org/10.1109/CVPR46437.2021.01556}
  {\path{doi:10.1109/CVPR46437.2021.01556}}.

\bibitem{prokudinDeepDirectionalStatistics2018}
S.~Prokudin, P.~Gehler, S.~Nowozin, Deep {{Directional Statistics}}: {{Pose
  Estimation}} with {{Uncertainty Quantification}}, in: V.~Ferrari, M.~Hebert,
  C.~Sminchisescu, Y.~Weiss (Eds.), Computer {{Vision}} -- {{ECCV}} 2018, Vol.
  11213, Springer International Publishing, Cham, 2018, pp. 542--559.
\newblock \href {https://doi.org/10.1007/978-3-030-01240-3_33}
  {\path{doi:10.1007/978-3-030-01240-3_33}}.

\bibitem{gilitschenskiDeepOrientationUncertainty2020}
I.~Gilitschenski, R.~Sahoo, W.~Schwarting, A.~Amini, S.~Karaman, D.~Rus, Deep
  {{Orientation Uncertainty Learning}} based on a {{Bingham Loss}} (2020).

\bibitem{xuRotatedObjectDetection2023}
H.~Xu, X.~Liu, Y.~Ma, Z.~Zhu, S.~Wang, C.~Yan, F.~Dai, Rotated {{Object
  Detection}} with {{Circular Gaussian Distribution}}, Electronics 12~(15)
  (2023) 3265.
\newblock \href {https://doi.org/10.3390/electronics12153265}
  {\path{doi:10.3390/electronics12153265}}.

\bibitem{yuPhaseShiftingCoderPredicting2023}
Y.~Yu, F.~Da, Phase-{{Shifting Coder}}: {{Predicting Accurate Orientation}} in
  {{Oriented Object Detection}} (Mar. 2023).
\newblock \href {http://arxiv.org/abs/2211.06368} {\path{arXiv:2211.06368}},
  \href {https://doi.org/10.48550/arXiv.2211.06368}
  {\path{doi:10.48550/arXiv.2211.06368}}.

\bibitem{xiaoTheoreticallyAchievingContinuous2024}
Z.-K. Xiao, G.-Y. Yang, X.~Yang, T.-J. Mu, J.~Yan, S.-M. Hu, Theoretically
  {{Achieving Continuous Representation}} of {{Oriented Bounding Boxes}}, in:
  2024 {{IEEE}}/{{CVF Conference}} on {{Computer Vision}} and {{Pattern
  Recognition}} ({{CVPR}}), IEEE, Seattle, WA, USA, 2024, pp. 16912--16922.
\newblock \href {https://doi.org/10.1109/CVPR52733.2024.01600}
  {\path{doi:10.1109/CVPR52733.2024.01600}}.

\bibitem{xuImageOrientationEstimation2024}
R.~Xu, Y.~Shi, Z.~Qi, Image {{Orientation Estimation Based On Deep Learning}} -
  {{A Survey}}, Procedia Computer Science 242 (2024) 1193--1197.
\newblock \href {https://doi.org/10.1016/j.procs.2024.08.176}
  {\path{doi:10.1016/j.procs.2024.08.176}}.

\bibitem{tsaiPreciseOrientationEstimation2024}
C.-Y. Tsai, W.-C. Lin, Precise {{Orientation Estimation}} for {{Rotated Object
  Detection Based}} on a {{Unit Vector Coding Approach}}, Electronics 13~(22)
  (2024) 4402.
\newblock \href {https://doi.org/10.3390/electronics13224402}
  {\path{doi:10.3390/electronics13224402}}.

\bibitem{nieDeepRotationCorrection2023}
L.~Nie, C.~Lin, K.~Liao, S.~Liu, Y.~Zhao, Deep {{Rotation Correction}} without
  {{Angle Prior}}, IEEE Trans. on Image Process. 32 (2023) 2879--2888.
\newblock \href {http://arxiv.org/abs/2207.03054} {\path{arXiv:2207.03054}},
  \href {https://doi.org/10.1109/TIP.2023.3275869}
  {\path{doi:10.1109/TIP.2023.3275869}}.

\bibitem{linMicrosoftCOCOCommon2015}
T.-Y. Lin, M.~Maire, S.~Belongie, L.~Bourdev, R.~Girshick, J.~Hays, P.~Perona,
  D.~Ramanan, C.~L. Zitnick, P.~Doll{\'a}r, Microsoft {{COCO}}: {{Common}}
  objects in context (2015).
\newblock \href {http://arxiv.org/abs/1405.0312} {\path{arXiv:1405.0312}}.

\bibitem{follmannRotationallyinvariantConvolutionModule2018}
P.~Follmann, T.~Bottger, A rotationally-invariant convolution module by feature
  map back-rotation, in: 2018 {{IEEE}} Winter Conference on Applications of
  Computer Vision ({{WACV}}), 2018, pp. 784--792.
\newblock \href {https://doi.org/10.1109/WACV.2018.00091}
  {\path{doi:10.1109/WACV.2018.00091}}.

\bibitem{russakovskyImageNetLargeScale2015}
O.~Russakovsky, J.~Deng, H.~Su, J.~Krause, S.~Satheesh, S.~Ma, Z.~Huang,
  A.~Karpathy, A.~Khosla, M.~Bernstein, A.~C. Berg, L.~{Fei-Fei}, {{ImageNet
  Large Scale Visual Recognition Challenge}} (Jan. 2015).
\newblock \href {http://arxiv.org/abs/1409.0575} {\path{arXiv:1409.0575}},
  \href {https://doi.org/10.48550/arXiv.1409.0575}
  {\path{doi:10.48550/arXiv.1409.0575}}.

\bibitem{dosovitskiyImageWorth16x162021}
A.~Dosovitskiy, L.~Beyer, A.~Kolesnikov, D.~Weissenborn, X.~Zhai,
  T.~Unterthiner, M.~Dehghani, M.~Minderer, G.~Heigold, S.~Gelly, J.~Uszkoreit,
  N.~Houlsby, An {{Image}} is {{Worth}} 16x16 {{Words}}: {{Transformers}} for
  {{Image Recognition}} at {{Scale}} (Jun. 2021).
\newblock \href {http://arxiv.org/abs/2010.11929} {\path{arXiv:2010.11929}},
  \href {https://doi.org/10.48550/arXiv.2010.11929}
  {\path{doi:10.48550/arXiv.2010.11929}}.

\bibitem{liuSwinTransformerHierarchical2021}
Z.~Liu, Y.~Lin, Y.~Cao, H.~Hu, Y.~Wei, Z.~Zhang, S.~Lin, B.~Guo, Swin
  {{Transformer}}: {{Hierarchical Vision Transformer}} using {{Shifted
  Windows}}, in: 2021 {{IEEE}}/{{CVF International Conference}} on {{Computer
  Vision}} ({{ICCV}}), IEEE, Montreal, QC, Canada, 2021, pp. 9992--10002.
\newblock \href {https://doi.org/10.1109/ICCV48922.2021.00986}
  {\path{doi:10.1109/ICCV48922.2021.00986}}.

\bibitem{wooConvNeXtV2Codesigning2023}
S.~Woo, S.~Debnath, R.~Hu, X.~Chen, Z.~Liu, I.~S. Kweon, S.~Xie, {{ConvNeXt
  V2}}: {{Co}}-designing and {{Scaling ConvNets}} with {{Masked Autoencoders}},
  in: 2023 {{IEEE}}/{{CVF Conference}} on {{Computer Vision}} and {{Pattern
  Recognition}} ({{CVPR}}), IEEE, Vancouver, BC, Canada, 2023, pp.
  16133--16142.
\newblock \href {https://doi.org/10.1109/CVPR52729.2023.01548}
  {\path{doi:10.1109/CVPR52729.2023.01548}}.

\bibitem{tanEfficientNetV2SmallerModels}
M.~Tan, Q.~V. Le, {{EfficientNetV2}}: {{Smaller Models}} and {{Faster
  Training}} (2021).

\bibitem{liuEfficientViTMemoryEfficient2023}
X.~Liu, H.~Peng, N.~Zheng, Y.~Yang, H.~Hu, Y.~Yuan, {{EfficientViT}}: {{Memory
  Efficient Vision Transformer}} with {{Cascaded Group Attention}}, in: 2023
  {{IEEE}}/{{CVF Conference}} on {{Computer Vision}} and {{Pattern
  Recognition}} ({{CVPR}}), IEEE, Vancouver, BC, Canada, 2023, pp.
  14420--14430.
\newblock \href {https://doi.org/10.1109/CVPR52729.2023.01386}
  {\path{doi:10.1109/CVPR52729.2023.01386}}.

\bibitem{maazEdgeNeXtEfficientlyAmalgamated2022}
M.~Maaz, A.~Shaker, H.~Cholakkal, S.~Khan, S.~W. Zamir, R.~M. Anwer, F.~S.
  Khan, {{EdgeNeXt}}: {{Efficiently Amalgamated CNN-Transformer Architecture}}
  for {{Mobile Vision Applications}} (Oct. 2022).
\newblock \href {http://arxiv.org/abs/2206.10589} {\path{arXiv:2206.10589}},
  \href {https://doi.org/10.48550/arXiv.2206.10589}
  {\path{doi:10.48550/arXiv.2206.10589}}.

\bibitem{yangFocalModulationNetworks}
J.~Yang, C.~Li, X.~Dai, J.~Gao, Focal {{Modulation Networks}} (2022).

\bibitem{yuMambaOutWeReally}
W.~Yu, X.~Wang, {{MambaOut}}: {{Do We Really Need Mamba}} for {{Vision}}?, in:
  Proceedings of the {{IEEE}}/{{CVF Conference}} on {{Computer Vision}} and
  {{Pattern Recognition}} ({{CVPR}}), Nashville, TN, USA, 2025, pp. 4484--4496.
\newblock \href {http://arxiv.org/abs/2405.07992} {\path{arXiv:2405.07992}}.

\bibitem{wightmanPyTorchImageModels2019}
R.~Wightman, {{PyTorch}} image models (2019).
\newblock \href {https://doi.org/10.5281/zenodo.4414861}
  {\path{doi:10.5281/zenodo.4414861}}.

\bibitem{loshchilovDecoupledWeightDecay2019}
I.~Loshchilov, F.~Hutter, Decoupled {{Weight Decay Regularization}} (Jan.
  2019).
\newblock \href {http://arxiv.org/abs/1711.05101} {\path{arXiv:1711.05101}},
  \href {https://doi.org/10.48550/arXiv.1711.05101}
  {\path{doi:10.48550/arXiv.1711.05101}}.

\bibitem{xuRethinkingBoundaryDiscontinuity2024}
H.~Xu, X.~Liu, H.~Xu, Y.~Ma, Z.~Zhu, C.~Yan, F.~Dai, Rethinking {{Boundary
  Discontinuity Problem}} for {{Oriented Object Detection}}, in: 2024
  {{IEEE}}/{{CVF Conference}} on {{Computer Vision}} and {{Pattern
  Recognition}} ({{CVPR}}), IEEE, Seattle, WA, USA, 2024, pp. 17406--17415.
\newblock \href {https://doi.org/10.1109/CVPR52733.2024.01648}
  {\path{doi:10.1109/CVPR52733.2024.01648}}.

\bibitem{yuBoundaryDiscontinuityAngle2024}
Y.~Yu, F.~Da, On {{Boundary Discontinuity}} in {{Angle Regression Based
  Arbitrary Oriented Object Detection}}, IEEE Trans. Pattern Anal. Mach.
  Intell. 46~(10) (2024) 6494--6508.
\newblock \href {https://doi.org/10.1109/TPAMI.2024.3378777}
  {\path{doi:10.1109/TPAMI.2024.3378777}}.

\bibitem{fanIdentificationBitmapCompression2003}
Z.~Fan, R.~L. {de Queiroz}, Identification of bitmap compression history:
  {{JPEG}} detection and quantizer estimation, IEEE Transactions on Image
  Processing 12~(2) (2003) 230--235.
\newblock \href {https://doi.org/10.1109/TIP.2002.807361}
  {\path{doi:10.1109/TIP.2002.807361}}.

\bibitem{popescuExposingDigitalForgeries2005}
A.~C. Popescu, H.~Farid, Exposing digital forgeries by detecting traces of
  resampling, IEEE Transactions on Signal Processing 53~(2) (2005) 758--767.
\newblock \href {https://doi.org/10.1109/TSP.2004.839932}
  {\path{doi:10.1109/TSP.2004.839932}}.

\bibitem{geirhosShortcutLearningDeep2020}
R.~Geirhos, J.-H. Jacobsen, C.~Michaelis, R.~Zemel, W.~Brendel, M.~Bethge,
  F.~A. Wichmann, Shortcut learning in deep neural networks, Nature Machine
  Intelligence 2~(11) (2020) 665--673.
\newblock \href {https://doi.org/10.1038/s42256-020-00257-z}
  {\path{doi:10.1038/s42256-020-00257-z}}.

\bibitem{doerschUnsupervisedVisualRepresentation2015}
C.~Doersch, A.~Gupta, A.~A. Efros, Unsupervised visual representation learning
  by context prediction, in: Proceedings of the {{IEEE International
  Conference}} on {{Computer Vision}} ({{ICCV}}), 2015, pp. 1422--1430.
\newblock \href {https://doi.org/10.1109/ICCV.2015.167}
  {\path{doi:10.1109/ICCV.2015.167}}.

\bibitem{gidarisUnsupervisedRepresentationLearning2018}
S.~Gidaris, P.~Singh, N.~Komodakis, Unsupervised representation learning by
  predicting image rotations, in: International Conference on Learning
  Representations ({{ICLR}}), 2018.

\bibitem{ortusaiRightWayUp2026}
{ORTUS AI}, {{RightWayUp}}: {{Image}} orientation models, benchmark notes and
  technical report, \url{https://github.com/ortusaitech/rightwayup} (2026).

\end{thebibliography}

\end{document}